\documentclass[10pt,twocolumn,letterpaper]{article}

\PassOptionsToPackage{pagebackref,breaklinks,colorlinks}{hyperref}

\usepackage[pagenumbers]{wacv} 

\usepackage[most]{tcolorbox}
\usepackage{orcidlink}
\usepackage{bm}
\usepackage{float}
\usepackage{tcolorbox}
\usepackage{tabularx}
\usepackage{multirow}
\usepackage{colortbl}
\usepackage{tabulary}
\usepackage{etoolbox}
\usepackage{pifont}
\usepackage{makecell}
\usepackage{graphicx}
\usepackage{adjustbox}
\usepackage{amsmath}
\usepackage{amssymb}
\usepackage{booktabs}
\usepackage{rotating}
\usepackage{tablefootnote}
\definecolor{amber}{rgb}{1.0, 0.75, 0.0}
\definecolor{LightCyan}{rgb}{0.8,1,1}
\definecolor{LightGreen}{rgb}{0.6,0.9,0.7}
\definecolor{royalblue(traditional)}{rgb}{0.0, 0.14, 0.4}
\definecolor{royalblue(web)}{rgb}{0.25, 0.41, 0.88}
\definecolor{darkgreen}{rgb}{0.0, 0.5, 0.0} 
\definecolor{darkred}{rgb}{0.5, 0.0, 0.0}
\definecolor{darkblue}{rgb}{0.0, 0.0, 0.8}
\usepackage{subcaption}
\usepackage{titletoc}
\usepackage{amsthm}
\newtheorem{theorem}{Theorem}[section]

\newtheorem{lemma}[theorem]{Lemma}
\newtheorem{proposition}[theorem]{Proposition}

\usepackage{adjustbox}

\newcommand{\checkm}{{\color{ForestGreen}\ding{51}}}
\newcommand{\crossm}{{\color{red}\ding{55}}}

\usepackage{tcolorbox}
\usepackage{xcolor}
\usepackage{pgfplots}
\usepackage{algorithm}
\usepackage{algpseudocode}
\usepackage{pgfplotstable}
\usepackage{filecontents}
\usepackage{mdframed}
\pgfplotsset{compat=1.17}
\usetikzlibrary{backgrounds,datavisualization.formats.functions}

\newcolumntype{Y}{>{\centering\arraybackslash}X} 
\newcolumntype{C}[1]{>{\centering\arraybackslash}m{#1}} 
\newcolumntype{L}[1]{>{\raggedright\arraybackslash}m{#1}} 
\newcolumntype{R}[1]{>{\raggedleft\a}}

\newcommand{\vect}[1]{\mathbf{#1}}

\definecolor{wacvblue}{rgb}{0.21,0.49,0.74}
\hypersetup{allcolors=wacvblue}

\def\wacvPaperID{548} 
\def\confName{WACV}
\def\confYear{2027}

\title{How (Mis)calibrated is your Federated CLIP and what to do about it?} 

\author{
\begin{tabular}{ccc}
Mainak Singha$^{1}$ &
Masih Aminbeidokhti$^{2}$ &
Paolo Casari$^{1}$ \\[0.3em]
Gianni Franchi$^{3}$ &
Elisa Ricci$^{1,4}$ &
Subhankar Roy$^{5}$
\end{tabular}
\\[0.8em]
\small
\begin{tabular}{cc}
$^{1}$ University of Trento, Italy &
$^{2}$ École de technologie supérieure, QC, Canada
\end{tabular}
\\
\small
\begin{tabular}{ccc}
$^{3}$ AMIAD, Pôle Recherche, Palaiseau, France &
$^{4}$ Fondazione Bruno Kessler, Italy &
$^{5}$ University of Bergamo, Italy
\end{tabular}
}

\begin{document}
\maketitle
\begin{abstract}
Vision-language models (VLMs) such as CLIP are increasingly adapted across decentralized data silos, yet the reliability of their predictions under federated learning (FL) remains largely unexplored. In this work, we present a systematic study of calibration in federated CLIP under non-IID client distributions. Our experiments reveal that widely used prompt-tuning methods consistently degrade calibration, often yielding substantially higher calibration error despite competitive recognition performance, while explicit training-time calibration regularizers provide only limited improvements. Motivated by these findings, we identify the choice of fine-tuning parameterization as a critical factor governing calibration and conduct a controlled comparison between prompt tuning and five backbone fine-tuning (BFT) strategies: AdaptFormer, LayerNorm, LoRA, VeRA, and DoRA. We find that BFT methods generally offer a more favorable accuracy-calibration trade-off than prompt tuning, although their benefits are not universal. Through extensive analysis, we show that calibration behavior is closely linked to the geometry of federated updates, residual parameterization, and the resulting client and logit drift. Across in-distribution, domain-generalization, and base-to-new evaluation settings, our results establish fine-tuning parameterization as a central design choice for building accurate and reliable federated CLIP models. Codes are available at \url{https://github.com/mainaksingha01/FL2oRA}.

\end{abstract}    
\section{Introduction}
\label{sec:intro}
Reliability is a critical consideration when deploying deep learning (DL) models in production scenarios. We aim to develop DL models that are not only highly accurate but also reliable. Reliability has many facets, including robustness to covariate shift \cite{csurka2017domain}, model calibration \cite{guo2017calibration}, and generalization to out-of-distribution data \cite{yang2024generalized}. Of particular interest, model calibration is crucial because it ensures that a model's predicted probabilities accurately reflect the true likelihood of correctness, enabling reliable decision-making \cite{pavlovic2025understanding}. As powerful DL models (\textit{e.g.}, CLIP \cite{radford2019language}) are increasingly being deployed in high-stake applications, ensuring proper calibration is crucial.

Model calibration is a widely studied topic in machine learning \cite{wang2023calibration,geng2024survey}. Prior works have shown that although vision-language models, such as CLIP, are reasonably well-calibrated for zero-shot image classification \cite{minderer2021revisiting}, fine-tuning them on downstream tasks often leads to miscalibration \cite{dor,wang2024open}. This is concerning because, due to its high generalizability, CLIP is often fine-tuned on small target datasets, which increases the risk of miscalibration. To alleviate miscalibration in downstream tasks, either \textit{post-hoc} calibration techniques (\textit{e.g.}, temperature scaling \cite{guo2017calibration}) or \textit{in-training} calibration techniques (\textit{e.g.}, DCA \cite{dca}, DOR \cite{dor}) have been used. However, fine-tuning is not only limited to the \textit{offline} setting, where a dataset is available in a centralized location \cite{coop,kgcoop,maple}, but also in federated learning (FL)~\cite{bai2024diprompt,promptfl,qiu2024federated,sun2024towards,fedotp,fedmvp,fedpgp,fedpha}, which is a more realistic setup. To our surprise, despite the practical relevance and progress in FL, \textit{no prior work has investigated how fine-tuning CLIP affects model calibration in the FL setting}.

\begin{figure*}[t]
    \centering
    \includegraphics[width=\linewidth]{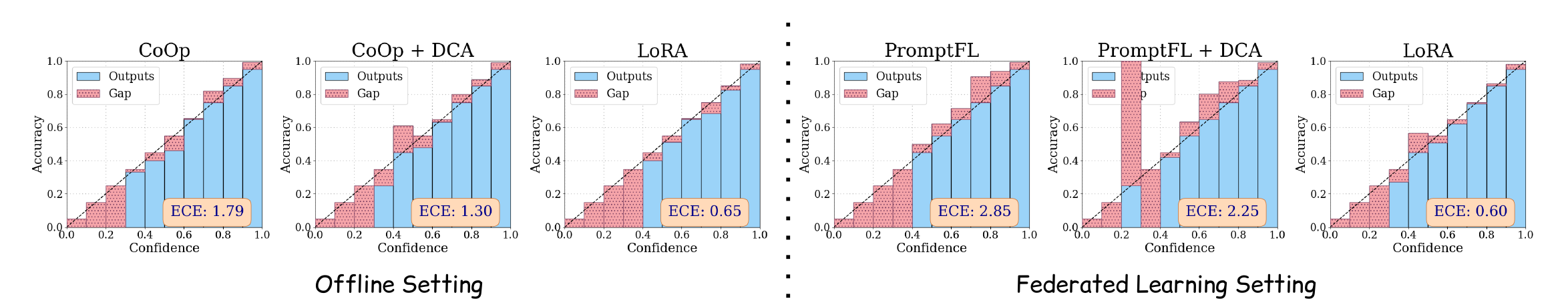}
\caption{\textbf{Comparison of reliability diagrams and calibration errors.} Expected Calibration Error (ECE $\downarrow$) \cite{guo2017calibration} is reported on the OxfordPets \cite{oxfordpets} dataset on a centralized training or offline setting ({left}) and a non-IID personalized federated learning setting ({right}). 
}
    \label{fig:ece_plots}
    \vspace{-15pt}
\end{figure*}

\begin{table}[tp]
    \centering
    \footnotesize
    \caption{\textbf{Overview of works focusing on model calibration.} While previous works have studied calibration of CLIP, no work has studied how fine-tuning CLIP in the federated learning (FL) setting affects calibration. Our work explores this direction.}
    \label{tab:benchmark}
    \resizebox{\linewidth}{!}{
        \begin{tabular}{lccccc}
            \toprule
            \textbf{Method} & \textbf{CLIP} & \textbf{Calibration} & \textbf{Fine-tune} & \textbf{Offline} & \textbf{FL} \\
            \midrule
            Minderer \textit{et al.}~\cite{minderer2021revisiting} & \checkm & \checkm & \crossm & \crossm & \crossm \\
            Wang \textit{et al.}~\cite{dor} & \checkm & \checkm & \checkm & \checkm & \crossm \\
            Chu \textit{et al.}~\cite{nucfl} & \crossm & \checkm & \checkm & \crossm & \checkm \\
            \midrule
            \textbf{Our work} & \checkm & \checkm & \checkm & \crossm & \checkm \\
            \bottomrule
        \end{tabular}
    }
\end{table}

In this work, we make an attempt to answer the question: \textit{\textbf{how much calibrated is CLIP when fine-tuned in a federated learning setup?}} Towards that goal, we first analyze if the trend of miscalibration, observed in the offline fine-tuning setting \cite{dor,wang2024open}, also holds true in the FL setting. As shown in Fig. \ref{fig:ece_plots}, CoOp \cite{coop}, a textual-prompt tuning approach, when used for fine-tuning CLIP in the FL setup (PromptFL \cite{promptfl}), it becomes more miscalibrated than its offline counterpart. Imposing regularization through in-training calibration (\textit{e.g.}, with DCA \cite{dca}) to both the offline and FL setting alleviates miscalibration by some extent. This observation conforms with the findings of \cite{nucfl} that studies model calibration in a non-CLIP FL setup. 

In more detail, as shown in Tab. \ref{tab:benchmark}, the work by Chu \textit{et al}. \cite{nucfl} only experimented with the classical ResNet family of models trained from scratch and did not explore a multimodal setting. Differently, in our study we are particularly interested in analyzing calibration error of fine-tuned CLIP models in the FL setup, which exhibits substantially different dynamics from standard Convnets \cite{minderer2021revisiting}. In Sec. \ref{sec:findings}, we dive deep into this issue, by investigating existing methods for improving calibration, and our empirical analyses show that \textit{improving calibration of CLIP-based methods in the FL setup is more challenging than the offline setting}.

Next, we show that calibration in federated CLIP depends not only on how the server aggregates client updates or whether an explicit calibration objective is used, but critically on which components of CLIP are fine-tuned. We first compare prompt-tuning methods previously proposed for FL (PromptFL \cite{promptfl}, 
FedPGP \cite{fedpgp}, FedOTP \cite{fedotp}, and FedPHA \cite{fedpha}) and confirm that they remain substantially miscalibrated even under training-time calibration. We then investigate, for the first time in federated CLIP, backbone fine-tuning (BFT), comprehensively comparing five representative families spanning bottleneck adapters, normalization tuning, and low-rank updates: including AdaptFormer \cite{adaptformer}, LayerNorm \cite{layernorm}, LoRA \cite{lora}, VeRA \cite{vera}, and DoRA \cite{dora}. Across three heterogeneous FL settings (\textit{i.e.}, in-distribution, domain-generalization, and base-to-new generalization), AdaptFormer, LoRA, and DoRA consistently provide the strongest accuracy-calibration trade-off, often matching or improving on calibration-regularized prompt tuning, while requiring no calibration objective at all. We further find that this behavior is governed primarily by the fine-tuning parameterization rather than the FL aggregation algorithm. We complement our empirical findings with a theoretical analysis investigating the behavior of backbone fine-tuning approaches in FL.

{\textbf{\emph{Contributions.}} We make the following contributions: ({{i}}) We are the first to study the impact of CLIP fine-tuning on calibration in FL. ({{ii}}) We systematically compare prompt tuning and backbone fine-tuning methods for federated CLIP, showing that backbone fine-tuning offers a better accuracy-calibration trade-off. ({{iii}}) We evaluate diverse FL scenarios and provide theoretical insights into the advantages of backbone fine-tuning over prompt tuning.
\section{Related Work}
\label{sec:formatting}
\noindent \textbf{\emph{Federated learning}} enables training across decentralized clients without sharing their local data. A major challenge is client heterogeneity, \textit{e.g.}, non-IID data, which can hurt performance. Existing methods mainly address this issue through improved aggregation or optimization, such as FedAvg \cite{fedavg}, FedProx \cite{fedprox}, and FedNova \cite{fednova}. While these approaches mostly target accuracy and convergence, we focus on a less explored question: how federated learning affects the calibration of Vision-Language Models (VLMs).

\noindent \textbf{\emph{Federated fine-tuning of CLIP.}}
When applied to large VLMs such as CLIP \cite{radford2019language}, FL introduces substantial computational and communication overhead due to the need to fine-tune millions of parameters. To mitigate this, prompt tuning~\cite{coop,kgcoop,vpt,maple} has emerged as an efficient alternative for adapting CLIP. Recent works extend prompt tuning to the federated setting~\cite{bai2024diprompt,promptfl,qiu2024federated,sun2024towards,fedotp,qiu2024federated,fedpgp,fedpha}. For example, PromptFL~\cite{promptfl} regularizes client-specific prompts, while FedOTP~\cite{fedotp} and FedTPG~\cite{qiu2024federated} improve communication efficiency and personalization via orthogonal or task-conditioned prompt generation. FedCLIP~\cite{fedclip} decouples global and local adaptation of the text and image encoders, and methods such as FedTPG~\cite{qiu2024federated} and FedMVP~\cite{fedmvp} dynamically generate prompts using auxiliary information (\textit{e.g.}, class names or visual attributes).

Our focus differs from these previous studies by investigating other PEFT approaches \cite{layernorm, lora, adaptformer, vera, dora} adapted to CLIP in federated settings, with particular emphasis on empirical performance and identifying the optimal accuracy-calibration trade-off.

\noindent \textbf{\emph{Model calibration}} \cite{pavlovic2025understanding} measures how well predicted confidence scores reflect true correctness likelihood. Calibration methods are generally divided into post-hoc and training-time approaches. Post-hoc methods, such as temperature scaling~\cite{guo2017calibration} and Dirichlet calibration~\cite{kull2017beta}, adjust output probabilities after training without modifying model parameters, offering simplicity but limited flexibility. In contrast, training-time methods, including label smoothing~\cite{muller2019does}, focal loss~\cite{lin2017focal}, DCA~\cite{dca}, and MDCA~\cite{hebbalaguppe2022stitch}, modify the objective to encourage well-calibrated predictions, often at the cost of additional hyperparameter tuning.

For VLMs, zero-shot CLIP is known to be well-calibrated~\cite{minderer2021revisiting}, but fine-tuning can degrade calibration, motivating approaches such as DOR~\cite{dor}, DAC~\cite{wang2024open}, and CalShift~\cite{khan2025confidence}. Most prior studies focus on offline settings. In FL, only a few works~\cite{peng2024fedcal,nucfl} address calibration, primarily with fully fine-tuned CNNs \cite{krizhevsky2012imagenet, he2016deep}. To our knowledge, calibration in CLIP-based FL remains unexplored.

\section{Preliminaries}
\label{sec:prelim}
In this section, we discuss the preliminaries related to FL,
federated fine-tuning of CLIP 
and model calibration. 
\vspace{+5pt}
\noindent \textbf{\emph{Federated learning.}} For the task of image classification, the FL scheme consists of training data distributed among $N$ clients and each client $n$ possess a dataset $\mathcal{D}_n = \{\vect{x}^{(i)}, y^{(i)}\}^{\mid \mathcal{D}_n \mid}_{i=1}$, where $\vect{x}^{(i)}$ denotes the $i$-th input image and $y^{(i)}$ denotes the class label. Ideally, we want to train a model by minimizing the empirical risk on the union of datasets of all the clients, $\mathcal{D} = \cup_{n \in N}\mathcal{D}_n$. However, due to the distributed nature of the datasets in the FL setting, we can only train a model $f_\theta$ on the local data of the client, \textit{i.e.}, conduct \textit{local training}, as:
\begin{equation}
    \label{eqn:ce_loss}
    \min_{\theta} \mathcal{L}(\theta) = \frac{1}{|\mathcal{D}_n|} \sum^{|\mathcal{D}_n|}_{i=1} \mathcal{L}_\text{ce}(f_\theta(\vect{x}^{(i)}), y^{(i)}),
\end{equation}
where $\theta$ are the weights of a client model and $\mathcal{L}_\text{ce}$ is cross-entropy loss.

To aggregate knowledge (or \textit{global aggregation}), the FL setting involves clients communicating the weights of the trained model $f_\theta$ to the server through multiple rounds of communication. In each round $t$, the server aggregates client weights via a weighted average, or, more concretely, using FedAvg \cite{fedavg} as:
\begin{equation}
    \label{eqn:fedavg}
    \theta_{t} = \sum^{N}_{n=1} \frac{|\mathcal{D}_n|}{|\mathcal{D}|} \theta^{(n)}_{t},
\end{equation}
where $\theta^{(n)}_{t}$ is the weights of the $n$-th client. After aggregation, the server sends back $\theta_t$ to the clients, which is used as a starting point for further local training.

In this work, we operate in a non-IID FL setup \cite{hsu2019measuring}, where each client $n$ can have varying number of classes, and the classes overlap among clients. This is more challenging than the prior CLIP-based FL literature \cite{qiu2024federated} that assumes the classes to be disjoint among the clients.

\vspace{+5pt}
\noindent \textbf{\emph{Federated fine-tuning of CLIP.}}
Textual prompt tuning, \textit{i.e.}, CoOp \cite{coop}, is a commonly used fine-tuning approach to improve the classification accuracy of CLIP on downstream tasks. For fine-tuning, it learns a set of prompts while keeping the weights of the pretrained CLIP image encoder $\mathcal{E}_v$ and text encoder $\mathcal{E}_t$ frozen. In detail, it replaces handcrafted prompts (\textit{e.g.}, ``\texttt{a photo of a }[CLS]'') at the input of the text encoder $\mathcal{E}_t$ with a set of lightweight learnable vectors (or tokens) $\mathcal{T} = \{\vect{v}_1, \vect{v}_2, \cdots, \vect{v}_M\}$, where $M$ denotes the length of the tokens. Text features are computed with the prompt $\mathbf{t} = \{\vect{v}_1, \vect{v}_2, \cdots, \vect{v}_M, [\text{CLS}]\}$, and the image features are computed from the image $\mathbf{x}$, using the respective encoders. The prompts $\mathcal{T}$ are learned using cross-entropy loss (Eq. (\ref{eqn:ce_loss})). 

Since the prompts $\mathcal{T}$ are lightweight and incur low communication overhead, it has become a \textbf{dominant approach} for fine-tuning CLIP in the FL setting \cite{promptfl,bai2024diprompt,fedclip}. Each client trains prompts $\mathcal{T}$ on the local client data and communicates them with the server. The server aggregates the prompts $\mathcal{T}$ from different clients using Eq. (\ref{eqn:fedavg}), and sends back the aggregated prompts to the clients. Prompt tuning-based FL methods differ mostly in how the prompts are designed, trained locally, and aggregated on the server. 

While prompt tuning is not the only method for fine-tuning CLIP in FL, it remains the most widely used. Related work has explored Adapters for FL of CLIP \cite{fedclip} and PEFT methods such as LoRA for fine-tuning LLMs/VLMs in multi-task and federated settings \cite{bohdal2025efficient,yang2025federated}.

\vspace{+5pt}
\noindent \textbf{\emph{Model calibration.}}
In addition to higher accuracy, we also want a model to be well-calibrated. That is, we want the predicted class probabilities of the model to be an accurate representation of the model's accuracy \cite{guo2017calibration}. Formally, a model is said to be well-calibrated (or \textit{confidence-calibrated}) if, for a confidence of $c$, the model is correct $c$ proportions of the time:
\begin{equation*}
    \mathbb{P}(y = \arg \max (\hat{p}(\vect{x})) \mid \max (\hat{p}(\vect{x})) = c) = c \quad \forall c \in [0, 1], 
\end{equation*}
where $(\vect{x}, y)$ is a datapoint, and $\hat{p}(\cdot)$ returns a discrete probability distribution over a set of $C$ classes. To quantify miscalibration the Expected Calibration Error (ECE) metric \cite{guo2017calibration} is used, which measures the difference between accuracy and confidence. To compute ECE for $|\mathcal{D}|$ samples in the dataset, we group the samples into $G$ equal-sized bins $\{b_1, b_2, \cdots b_G\}$ based on the prediction confidence of the model, and take a weighted average of the absolute difference between the average accuracy and average confidence:
\begin{equation*}
    \text{ECE} = \sum^{G}_{g=1} \frac{\mid b_g \mid}{|\mathcal{D}|} \mid \text{acc}(b_g) - \text{conf}(b_g) \mid,
\end{equation*}
where $\text{acc}(b_g)$ and $\text{conf}(g_g)$ are the average accuracy and average confidence for the samples falling in the bin $b_g$, respectively. A well-calibrated model will have a low ECE.

In centralized (offline) settings, prior work showed that while zero-shot CLIP is well-calibrated in few-shot classification \cite{minderer2021revisiting}, fine-tuning can degrade calibration \cite{wang2024open,dor}. To mitigate this, methods introduce regularization (\textit{e.g.}, constraining text encoder divergence \cite{dor}) during fine-tuning. However, the impact of fine-tuning CLIP on calibration in FL remains unexplored, which is the goal of our work.

\section{Experimental setup}
\label{sec:exp_setup}

\noindent\textbf{\emph{Benchmarks and settings.}} For our evaluation, we use 10 benchmarks: CIFAR-10, CIFAR-100 \cite{cifar}, OfficeHome \cite{officehome}, PACS \cite{pacs}, VLCS \cite{vlcs}, Food101 \cite{food}, DTD \cite{dtd}, Caltech101 \cite{caltech}, Flowers102 \cite{oxfordflowers}, and OxfordPets \cite{oxfordpets}. We follow the non-IID FL setting, where the clients have training data of variable number of classes and the classes can be shared among the clients, \textit{i.e.} overlapping classes. Following previous work \cite{cao2023knowledge,shamsian2021personalized}, we simulate data heterogeneity by partitioning the data among the clients by setting the parameter $\beta=0.5$ of a symmetric Dirichlet distribution.

We organize the benchmarks to simulate three heterogeneous FL scenarios: (i) an \textit{in-distribution} setting where a single dataset is split across clients, (ii) a \textit{domain generalization} setting where different domains are assigned to different clients, and (iii) a \textit{base-to-new generalization} setting where some classes are held out during training. Additional details are provided in the \texttt{Supplementary}. 

\noindent\textbf{\emph{FL Baselines.}} We evaluate several prompt tuning methods for CLIP-based image classification which have been previously proposed for the FL setting: PromptFL \cite{promptfl}, which extends CoOp \cite{coop} to FL, FedPGP \cite{fedpgp}, FedOTP \cite{fedotp}, and FedPHA \cite{fedpha}. For off-the-shelf calibration, we use temperature scaling \cite{guo2017calibration} (post-hoc method) and two training-based methods: DCA \cite{dca} and MDCA \cite{hebbalaguppe2022stitch}.

\noindent\textbf{\emph{Implementation Details.}} We use a frozen CLIP ViT-B/16 backbone. We set the Dirichlet parameter to $\beta=0.5$, run 50 communication rounds with 1 local epoch, and use 100\% participation rate (10\% for CIFAR-10/100). For base-to-new generalization, we use 16 shots per class across 10 clients; for domain generalization, two clients per domain; and for in-distribution settings, 100 clients per dataset. We use the standard optimizer SGD \cite{sgd}, a learning rate of $1\times10^{-3}$, warm-up $1\times10^{-5}$, and batch size 32.

\noindent\textbf{\emph{Evaluation metrics.}} To compare methods, we compute: classification accuracy (Acc $\uparrow$), Expected Calibration Error (ECE $\downarrow$) \cite{guo2017calibration}, Maximum Calibration Error (MCE $\downarrow$) \cite{guo2017calibration}, and Adaptive Calibration Error (ACE $\downarrow$) \cite{nixon2019measuring}. We report performances in the FL setting \cite{fedotp,fedpgp,fedpha}, where the metrics are computed at each client and then averaged.

\begin{table*}[!t]
    \centering
    \caption{\textbf{Comparison of offline versus FL fine-tuning methods}. Results have been reported on the \textit{in-distribution} setting using datasets: CIFAR-10 and CIFAR-100. Best number is in \textbf{bold}.}
    \scalebox{0.70}{
    \begin{tabular}{l|cccc|cccc|cccc}
        \toprule
        
        \multirow{3}{*}{\textbf{Method}}
        & \multicolumn{8}{c|}{\textbf{Offline setting}}
        & \multicolumn{4}{c}{\textbf{FL setting}} \\
        
        \cmidrule(lr){2-9}
        \cmidrule(lr){10-13}
        
        & \multicolumn{4}{c|}{\textbf{w/o calib}}
        & \multicolumn{4}{c|}{\textbf{w/ calib: DCA \cite{dca}}}
        & \multicolumn{4}{c}{\textbf{w/o calib}} \\
        
        \cmidrule(lr){2-5}
        \cmidrule(lr){6-9}
        \cmidrule(lr){10-13}
        
        & Acc. $\uparrow$
        & ECE $\downarrow$
        & MCE $\downarrow$
        & ACE $\downarrow$
        & Acc. $\uparrow$
        & ECE $\downarrow$
        & MCE $\downarrow$
        & ACE $\downarrow$
        & Acc. $\uparrow$
        & ECE $\downarrow$
        & MCE $\downarrow$
        & ACE $\downarrow$ \\
        
        \midrule
        
        \rowcolor{gray!30}
        \multicolumn{13}{c}{\textbf{CIFAR-10}} \\
        
        CoOp \cite{coop}
        & 93.99 & \textbf{1.00} & 0.56 & 0.89
        & 93.89 & 1.20 & 0.62 & \textbf{0.87}
        & 92.42 & 4.94 & 1.64 & 3.94 \\
        
        KgCoOp \cite{kgcoop}
        & 92.17 & 2.70 & 0.61 & 2.70
        & 92.37 & 2.21 & \textbf{0.46} & 2.21
        & 90.48 & 6.53 & 1.88 & 5.79 \\

        VPT \cite{vpt}
        & 96.93 & 1.23 & 0.64 & 0.90
        & \textbf{96.98} & 1.45 & 0.75 & 0.94
        & 95.29 & 3.53 & 1.22 & 2.52 \\
        
        \midrule
        
        \rowcolor{gray!30}
        \multicolumn{13}{c}{\textbf{CIFAR-100}} \\
        
        CoOp \cite{coop}
        & 77.86 & 1.93 & 0.57 & 1.77
        & 77.85 & 1.83 & 0.73 & 1.71
        & 72.43 & 9.27 & 2.43 & 8.66 \\
        
        KgCoOp \cite{kgcoop}
        & 74.73 & 7.83 & 1.13 & 7.83
        & 74.82 & 5.30 & 0.80 & 5.30
        & 70.68 & 11.20 & 2.95 & 11.05 \\

        VPT \cite{vpt}
        & \textbf{80.86} & \textbf{1.45} & 0.41 & \textbf{1.35}
        & 80.82 & 1.47 & \textbf{0.39} & 1.46
        & 77.83 & 7.94 & 2.39 & 8.00 \\
        
        \bottomrule
    \end{tabular}
    }
    \label{tab:cifar_offline_horizontal}
\end{table*}

\section{Empirical Findings}
\label{sec:findings}
In this section, we first examine how federated fine-tuning with prompt tuning methods affects the calibration of CLIP-based models (Sec. \ref{sec:calib_baselines}). We then empirically investigate strategies for improving calibration of CLIP (Sec. \ref{sec:improving}). In particular, we study whether existing calibration techniques and federated optimization strategies can alleviate miscalibration, and evaluate the accuracy-calibration trade-off of alternative PEFT methods across three experimental settings. Finally, we assess whether post-hoc temperature scaling provides additional calibration benefits for PEFT methods (Sec. \ref{sec:posthoc_calibration}), and examine the robustness of the performance gains by varying communication budgets and client data heterogeneity (Sec. \ref{sec:fl_robustness}).
\subsection{How does FL affect calibration when using prompt tuning methods?}
\label{sec:calib_baselines}
We begin by comparing the calibration error of textual prompt tuning-based methods (\textit{e.g.}, CoOp \cite{coop} and KgCoOp \cite{kgcoop}), as well as Visual Prompt Tuning (VPT) \cite{vpt}, in both the offline and FL settings. Our goal is to understand whether the miscalibration of prompt-based methods, observed in \cite{dor} in the offline CLIP fine-tuning setting, also holds true in the FL setting. \cref{tab:cifar_offline_horizontal} reports the performance of methods: ({i}) in the \textit{offline} setting \textit{without} calibration, ({ii}) in the \textit{offline} setting \textit{with} calibration (\textit{i.e.}, with DCA \cite{dca}), and ({iii}) in the \textit{FL} setting \textit{without} calibration.  The FL experiments were performed in the \textit{in-distribution} setting. 

From \cref{tab:cifar_offline_horizontal} we observe that, on average across metrics and datasets, prompt tuning methods in the offline setting exhibit lower calibration error when combined with a calibration technique. This aligns with prior empirical findings \cite{dor}. 
However, a previously unexplored question is: \textit{to what extent does fine-tuning CLIP in a FL setting worsen calibration?} As shown in the last column of Tab.~\ref{tab:cifar_offline_horizontal}, the same methods exhibit substantially higher calibration error in FL across all metrics. For example, on CIFAR-100, the ECE of CoOp increases from 1.93\% in the offline setting to 9.27\% in FL, corresponding to a 4.8$\times$ increase. While prior work \cite{nucfl} reports similar trends, their findings are limited to ResNet architectures trained from scratch. In contrast, we study CLIP fine-tuning in FL, which is particularly relevant given its widespread use for many downstream tasks.

In summary, the results in Tab.~\ref{tab:cifar_offline_horizontal} indicate that \textbf{prompt tuning (PT) methods, predominant in the CLIP-based FL setting, consistently introduce miscalibration}. Therefore, assessing CLIP-based FL methods solely based on accuracy, as done in several recent FL works \cite{fedotp,fedpha,fedpgp}, provides an incomplete evaluation of their performance.

\begin{table*}[!t]
    \centering
    \caption{\textbf{Comparison of FL methods on the \textit{in-distribution} setting} using datasets: CIFAR-10 and CIFAR-100. Best number is in \textbf{bold}, and second best is \underline{underlined}. Legends: \colorbox{amber!13}{prompt tuning methods}, and \colorbox{darkgreen!13}{backbone fine-tuning methods}.}
    \scalebox{0.70}{
    \begin{tabular}{l|cccc|cccc|cccc}
         \toprule
         \textbf{Method} 
         & \multicolumn{4}{c|}{\textbf{\makecell{FL setting \\(w/o calib)}}} 
         & \multicolumn{4}{c|}{\textbf{\makecell{FL setting \\(w/ calib: DCA \cite{dca})}}} 
         & \multicolumn{4}{c}{\textbf{\makecell{FL setting \\(w/ calib: MDCA \cite{hebbalaguppe2022stitch})}}} \\
         
         & Acc. $\uparrow$ & ECE $\downarrow$ & MCE $\downarrow$ & ACE $\downarrow$
         & Acc. $\uparrow$ & ECE $\downarrow$ & MCE $\downarrow$ & ACE $\downarrow$
         & Acc. $\uparrow$ & ECE $\downarrow$ & MCE $\downarrow$ & ACE $\downarrow$ \\
         \midrule
         
         \rowcolor{gray!30}
         \multicolumn{13}{c}{CIFAR-10} \\

         \rowcolor{amber!13} 
         PromptFL \cite{promptfl}
         & 92.42 & 4.94 & 1.64 & 3.94
         & 92.12 & 4.98 & 1.73 & 3.92
         & 92.31 & 5.07 & 1.67 & 3.95 \\

         \rowcolor{amber!13}
         FedPGP \cite{fedpgp}
         & 91.28 & 5.89 & 1.78 & 5.17
         & 91.54 & 5.65 & 1.75 & 4.69
         & 90.98 & 5.68 & 1.79 & 5.06 \\

         \rowcolor{amber!13}
         FedOTP \cite{fedotp}
         & 94.38 & 6.03 & 1.74 & 5.37
         & 94.30 & 5.41 & 1.59 & 4.53
         & 94.24 & 5.83 & 1.77 & 5.17 \\

         \rowcolor{amber!13}
         FedPHA \cite{fedpha}
         & 93.72 & 4.59 & 1.47 & 3.52
         & 93.53 & 4.81 & 1.46 & 3.41
         & 93.55 & 4.78 & 1.53 & 3.38 \\
         
         \midrule

         \rowcolor{darkgreen!13}AdaptFormer \cite{adaptformer}
         & 96.80 & \underline{3.00} & 1.16 & \underline{1.93}
         & 96.75 & \textbf{2.95} & \underline{1.12} & \underline{1.87}
         & 96.73 & \textbf{2.98} & \underline{1.13} & \underline{1.89} \\
         
         \rowcolor{darkgreen!13}LayerNorm \cite{layernorm}
         & 96.03 & 3.49 & 1.30 & 2.21
         & 95.98 & 3.37 & 1.24 & 2.10
         & 95.95 & 3.42 & 1.27 & 2.14 \\

         \rowcolor{darkgreen!13}LoRA \cite{lora}
         & \underline{97.04} & \textbf{2.99} & \underline{1.12} & 2.06
         & \underline{97.00} & \underline{2.99} & 1.15 & \underline{1.87}
         & \underline{96.92} & \underline{3.03} & 1.15 & 1.96 \\
         
         \rowcolor{darkgreen!13}
         VeRA \cite{vera}
         & 88.93 & 8.30 & 2.25 & 7.82
         & 89.02 & 8.08 & 2.20 & 7.59
         & 88.87 & 8.19 & 2.23 & 7.70 \\
         \rowcolor{darkgreen!13}
         DoRA \cite{dora}
         & \textbf{97.19} & 3.05 & \textbf{1.08} & \textbf{1.85}
         & \textbf{97.14} & 3.01 & \textbf{1.07} & \textbf{1.80}
         & \textbf{97.11} & 3.04 & \textbf{1.08} & \textbf{1.82} \\
         
         \midrule
         
         \rowcolor{gray!30}
         \multicolumn{13}{c}{CIFAR-100} \\

         \rowcolor{amber!13}
         PromptFL \cite{promptfl}
         & 72.43 & 9.27 & 2.43 & 8.66
         & 72.07 & 8.92 & 2.34 & 8.77
         & 72.02 & 9.18 & 2.45 & 8.85 \\

         \rowcolor{amber!13}
         FedPGP \cite{fedpgp}
         & 71.13 & 9.91 & 2.59 & 9.78
         & 71.36 & 10.38 & 2.73 & 10.14
         & 71.88 & 9.82 & 2.66 & 9.65 \\

         \rowcolor{amber!13}
         FedOTP \cite{fedotp}
         & 74.81 & 14.44 & 3.63 & 14.30
         & 74.63 & 12.67 & 3.27 & 12.52
         & 74.73 & 14.42 & 3.60 & 14.45 \\

         \rowcolor{amber!13}
         FedPHA \cite{fedpha}
         & 74.23 & 8.91 & 2.33 & 8.61
         & 75.30 & 8.06 & 2.20 & 7.63
         & 75.34 & 8.65 & 2.39 & 8.11 \\
         
         \midrule
         
         \rowcolor{darkgreen!13}
         AdaptFormer \cite{adaptformer}
         & 82.64 & 6.98 & 2.10 & 6.48
         & 82.58 & 6.77 & \underline{2.05} & 6.25
         & 82.52 & 6.87 & 2.07 & 6.34 \\
         
         \rowcolor{darkgreen!13}
         LayerNorm \cite{layernorm}
         & 81.33 & 7.41 & 2.25 & 6.92
         & 81.27 & 7.18 & 2.18 & 6.70
         & 81.20 & 7.30 & 2.22 & 6.80 \\

         \rowcolor{darkgreen!13}
         LoRA \cite{lora}
         & \textbf{83.70} & \textbf{6.70} & \underline{2.05} & \textbf{6.09}
         & \textbf{83.65} & \textbf{6.64} & 2.12 & \underline{6.20}
         & \textbf{83.73} & \underline{6.82} & \textbf{2.01} & \textbf{6.13} \\
         
         \rowcolor{darkgreen!13}
         VeRA \cite{vera}
         & 72.84 & 10.57 & 2.82 & 10.18
         & 72.91 & 10.15 & 2.73 & 9.78
         & 72.79 & 10.37 & 2.79 & 9.99 \\

         \rowcolor{darkgreen!13}
         DoRA \cite{dora}
         & \underline{83.18} & \underline{6.88} & \textbf{2.03} & \underline{6.31}
         & \underline{83.12} & \underline{6.72} & \textbf{2.00} & \textbf{6.16}
         & \underline{83.09} & \textbf{6.81} & \underline{2.02} & \underline{6.22} \\
         
         \bottomrule
    \end{tabular}
    }
    \label{tab:cifar_indistribution_horizontal}
\end{table*}

\subsection{Can we improve calibration?}
\label{sec:improving}
Given that prompt tuning methods worsen calibration in FL, a natural question is \textbf{{whether calibration can be meaningfully improved and how.}} 

\vspace{+5pt}
\noindent\textbf{\emph{Effect of calibration techniques.}}
Following \cite{dor}, a seemingly apparent solution to improve calibration is to impose calibration regularization during fine-tuning. We applied such techniques (\textit{e.g.}, with DCA \cite{dca} or MDCA \cite{hebbalaguppe2022stitch}) to state-of-the-art FL baselines methods based on prompt tuning in three settings: \textit{in-distribution} (Tab. \ref{tab:cifar_indistribution_horizontal}), \textit{domain generalization} (Tab. \ref{tab:dg_horizontal}), and \textit{base-to-new generalization} (Tab. \ref{tab:bton_horizontal}). The purpose is to compare performance with and without calibration regularization. In the \textit{in-distribution} setting (\cref{tab:cifar_indistribution_horizontal}), applying DCA or MDCA to FL prompt tuning methods yields only modest improvements in ECE and ACE for certain configurations (\textit{e.g.}, PromptFL and FedPHA on CIFAR-100), while many methods remain substantially miscalibrated. For example, on CIFAR-100, FedOTP still exhibits ECE above 12\% with DCA and above 14\% with MDCA. A similar pattern appears on \textit{domain generalization} and \textit{base-to-new generalization} (Tabs.~\ref{tab:dg_horizontal} and~\ref{tab:bton_horizontal}). DCA and MDCA slightly improve calibration on FedPGP and FedPHA in some cases, yet other methods (again notably FedOTP) stay severely miscalibrated, with ECE often exceeding 7--9\%. Overall, we observe that \textbf{in-training calibration in prompt tuning methods for FL rarely produces large, systematic improvements and sometimes even makes it more miscalibrated}.

\begin{table*}[!t]
    \centering
    \caption{\textbf{Comparison of FL methods on the \textit{domain generalization} setting} using the datasets: OfficeHome, PACS, and VLCS. Best number is in \textbf{bold}, and second best is \underline{underlined}. Legends: \colorbox{amber!13}{prompt tuning methods}, and \colorbox{darkgreen!13}{backbone fine-tuning methods}.}
    \scalebox{0.70}{
    \begin{tabular}{l|cccc|cccc|cccc}
         \toprule
         \textbf{Method} 
         & \multicolumn{4}{c|}{\textbf{\makecell{FL setting \\(w/o calib)}}} 
         & \multicolumn{4}{c|}{\textbf{\makecell{FL setting \\(w/ calib: DCA \cite{dca})}}} 
         & \multicolumn{4}{c}{\textbf{\makecell{FL setting \\(w/ calib: MDCA \cite{hebbalaguppe2022stitch})}}} \\
         
         & Acc. $\uparrow$ & ECE $\downarrow$ & MCE $\downarrow$ & ACE $\downarrow$
         & Acc. $\uparrow$ & ECE $\downarrow$ & MCE $\downarrow$ & ACE $\downarrow$
         & Acc. $\uparrow$ & ECE $\downarrow$ & MCE $\downarrow$ & ACE $\downarrow$ \\
         \midrule
         
         \rowcolor{amber!13}
         PromptFL \cite{promptfl} 
         & 91.25 & 2.87 & 1.21 & 2.78
         & 91.33 & 3.07 & 1.22 & 2.87
         & 91.32 & 2.91 & 1.21 & 2.77 \\

         \rowcolor{amber!13}
         FedPGP \cite{fedpgp} 
         & 92.72 & 2.67 & 0.89 & 2.24
         & \underline{93.93} & 2.28 & 0.68 & 2.06
         & 94.04 & 2.29 & 0.90 & 1.90 \\

         \rowcolor{amber!13}
         FedOTP \cite{fedotp} 
         & 91.07 & 9.34 & 2.36 & 9.21
         & 91.53 & 7.37 & 1.90 & 7.27
         & 91.88 & 9.26 & 2.40 & 9.24 \\

         \rowcolor{amber!13}
         FedPHA \cite{fedpha} 
         & 92.87 & 2.29 & 0.81 & 1.96
         & \underline{93.93} & 2.17 & 0.64 & 2.03
         & \underline{94.06} & 2.23 & 0.94 & 1.84 \\

         \midrule
         
         \rowcolor{darkgreen!13}AdaptFormer \cite{adaptformer} 
         & \underline{93.02} & \underline{1.98} & \underline{0.70} & \underline{1.72}
         & 93.72 & \underline{1.74} & \underline{0.52} & \underline{1.57}
         & 93.93 & \underline{1.66} & \underline{0.57} & \underline{1.34} \\

         \rowcolor{darkgreen!13}LayerNorm \cite{layernorm} & 90.65 & 2.95 & 0.95 & 2.54 & 91.10 & 2.67 & 0.84 & 2.28 & 91.24 & 2.54 & 0.80 & 2.16 \\
         
         \rowcolor{darkgreen!13}LoRA \cite{lora} 
         & \textbf{93.11} & 2.37 & 0.79 & 1.83
         & 93.13 & 2.10 & 0.72 & 1.95
         & 93.07 & 2.24 & 0.77 & 1.64 \\

        \rowcolor{darkgreen!13}VeRA \cite{vera} 
         & 91.03 & 3.29 & 1.10 & 2.94
         & 91.73 & 2.90 & 0.96 & 2.60
         & 91.84 & 2.96 & 1.00 & 2.65 \\
         
         \rowcolor{darkgreen!13}DoRA \cite{dora} 
         & \textbf{93.11} & \textbf{1.88} & \textbf{0.65} & \textbf{1.63}
         & \textbf{93.95} & \textbf{1.59} & \textbf{0.48} & \textbf{1.47}
         & \textbf{94.25} & \textbf{1.47} & \textbf{0.50} & \textbf{1.24} \\
         
         \bottomrule
    \end{tabular}
    }
    \label{tab:dg_horizontal}
\end{table*}
\begin{table*}[!t]
    \centering
    \caption{\textbf{Comparison of FL methods on the \textit{base-to-new generalization} setting} using the datasets: Food101, DTD, Caltech101, Flowers102, and OxfordPets. Best number is in \textbf{bold}, and second best is \underline{underlined}. Legends: \colorbox{amber!13}{prompt tuning methods}, and \colorbox{darkgreen!13}{backbone fine-tuning methods}.}
    \scalebox{0.70}{
    \begin{tabular}{l|cccc|cccc|cccc}
         \toprule
         \textbf{Method}
         & \multicolumn{4}{c|}{\textbf{\makecell{FL setting \\(w/o calib)}}}
         & \multicolumn{4}{c|}{\textbf{\makecell{FL setting \\(w/ calib: DCA \cite{dca})}}}
         & \multicolumn{4}{c}{\textbf{\makecell{FL setting \\(w/ calib: MDCA \cite{hebbalaguppe2022stitch})}}} \\

         & Acc. $\uparrow$ & ECE $\downarrow$ & MCE $\downarrow$ & ACE $\downarrow$
         & Acc. $\uparrow$ & ECE $\downarrow$ & MCE $\downarrow$ & ACE $\downarrow$
         & Acc. $\uparrow$ & ECE $\downarrow$ & MCE $\downarrow$ & ACE $\downarrow$ \\
         \midrule

         \rowcolor{amber!13}
         PromptFL \cite{promptfl}
         & 82.98 & 4.11 & \textbf{1.32} & 3.82
         & 84.25 & \textbf{3.97} & \textbf{1.30} & 3.62
         & 84.62 & \textbf{4.06} & \textbf{1.23} & 3.73 \\

         \rowcolor{amber!13}
         FedPGP \cite{fedpgp}
         & 81.10 & 6.17 & 2.07 & 5.97
         & 80.90 & 5.82 & 1.82 & 5.69
         & 81.51 & 4.93 & 1.65 & 4.80 \\

         \rowcolor{amber!13}
         FedOTP \cite{fedotp}
         & 29.52 & 21.88 & 5.03 & 21.90
         & 28.64 & 26.58 & 7.34 & 26.63
         & 28.53 & 25.35 & 6.25 & 25.32 \\

         \rowcolor{amber!13}
         FedPHA \cite{fedpha}
         & 54.97 & 16.03 & 4.64 & 15.69
         & 54.56 & 17.93 & 5.53 & 17.38
         & 52.33 & 20.55 & 6.63 & 19.86 \\

         \midrule

         \rowcolor{darkgreen!13}AdaptFormer \cite{adaptformer}
         & 86.14 & 3.97 & 1.43 & 3.24
         & 85.50 & 4.23 & 1.62 & 3.55
         & 86.10 & 4.14 & 1.58 & 3.60 \\

         \rowcolor{darkgreen!13}LayerNorm \cite{layernorm}
         & 85.13 & 5.17 & 2.02 & 4.37
         & 84.82 & 5.66 & 2.14 & 4.75
         & 85.30 & 5.36 & 2.05 & 4.64 \\

         \rowcolor{darkgreen!13}LoRA \cite{lora}
         & \underline{86.25} & \underline{3.91} & \underline{1.41} & \underline{3.20}
         & \textbf{86.08} & \underline{4.21} & \underline{1.61} & \textbf{3.32}
         & \underline{86.34} & \underline{4.10} & 1.56 & \textbf{3.35} \\


         \rowcolor{darkgreen!13}VeRA \cite{vera}
         & 82.20 & 5.61 & 1.83 & 5.26
         & 82.49 & 5.02 & 1.54 & 4.68
         & 82.78 & 4.75 & \underline{1.47} & 4.43 \\

         \rowcolor{darkgreen!13}DoRA \cite{dora}
         & \textbf{86.40} & \textbf{3.70} & 1.46 & \textbf{3.00}
         & \underline{85.90} & 4.25 & 1.64 & \underline{3.45}
         & \textbf{86.50} & 4.13 & 1.58 & \underline{3.50} \\

         \bottomrule
    \end{tabular}
    }
    \label{tab:bton_horizontal}
\end{table*}

\vspace{+5pt}
\begin{table*}[!t]
    \centering
    \caption{\textbf{Comparison of FL optimization algorithms on backbone fine-tuning methods}. Results have been reported on the \textit{in-distribution} setting \textit{without} applying any calibration using CIFAR-100 dataset. Best number is in \textbf{bold}, and second best is \underline{underlined}.}
    \scalebox{0.7}{
    \begin{tabular}{l|cccc|cccc|cccc|cccc}
         \toprule
         \textbf{Method}
         & \multicolumn{4}{c|}{\textbf{FedAvg} \cite{fedavg}}
         & \multicolumn{4}{c|}{\textbf{FedProx} \cite{fedprox}}
         & \multicolumn{4}{c|}{\textbf{FedDyn} \cite{feddyn}}
         & \multicolumn{4}{c}{\textbf{FedNova} \cite{fednova}} \\
         
         & Acc. $\uparrow$ & ECE $\downarrow$ & MCE $\downarrow$ & ACE $\downarrow$
         & Acc. $\uparrow$ & ECE $\downarrow$ & MCE $\downarrow$ & ACE $\downarrow$
         & Acc. $\uparrow$ & ECE $\downarrow$ & MCE $\downarrow$ & ACE $\downarrow$
         & Acc. $\uparrow$ & ECE $\downarrow$ & MCE $\downarrow$ & ACE $\downarrow$ \\
         \midrule

         AdaptFormer \cite{adaptformer}
         & 82.95 & 7.05 & 2.16 & 6.42
         & 83.02 & 7.18 & 2.22 & 6.55
         & 82.85 & 6.98 & 2.18 & 6.48
         & 82.18 & 7.45 & 2.71 & 7.02 \\

         LayerNorm \cite{layernorm}
         & 81.33 & 7.41 & 2.25 & 6.92
         & 81.95 & 7.62 & 2.31 & 7.10
         & 81.70 & 7.35 & 2.27 & 6.98
         & 80.95 & 7.88 & 2.84 & 7.52 \\

         LoRA \cite{lora}
         & \textbf{83.70} & 6.70 & \textbf{2.05} & 6.09
         & \underline{83.55} & 6.86 & 2.14 & 6.15
         & 83.42 & 6.79 & 2.15 & 6.15
         & 82.89 & 7.03 & 2.56 & 6.67 \\

         VeRA \cite{vera}
         & 80.95 & 7.72 & 2.29 & 7.10
         & 81.60 & 7.95 & 2.40 & 7.22
         & 81.45 & 7.65 & 2.32 & 7.14
         & 80.72 & 8.10 & 2.88 & 7.62 \\

         DoRA \cite{dora}
         & 83.42 & \textbf{6.58} & 2.10 & 6.18
         & 83.38 & 6.72 & 2.11 & \underline{6.08}
         & 83.20 & \underline{6.65} & \underline{2.09} & \textbf{6.02}
         & 82.66 & 6.91 & 2.48 & 6.54 \\

         \bottomrule
    \end{tabular}
    }
    \label{tab:fl_algo}
\end{table*}

\noindent\textbf{\emph{Improving calibration through backbone fine-tuning.}}
The limited gains from applying DCA and MDCA to PT methods suggest that their miscalibration cannot be resolved reliably through the calibration objective alone. We therefore ask whether calibration depends on \textit{what} is fine-tuned and \textit{how} the update is parameterized. To this end, we compare prompt tuning methods with several PEFT approaches that update the intermediate representation of the backbone, namely: AdaptFormer \cite{adaptformer}, LayerNorm \cite{layernorm}, LoRA \cite{lora}, VeRA \cite{vera}, and DoRA \cite{dora}. These methods span bottleneck adapters, normalization tuning, and low-rank updates under the same federated protocol. We refer to these methods as \textbf{backbone fine-tuning (BFT) methods} as they update the intermediate representations of the encoders, unlike PT that modifies just the text classifier. Implementation details are provided in the \texttt{Supplementary}.

Across the three FL settings we consider (in Tabs.~\ref{tab:cifar_indistribution_horizontal}, \ref{tab:dg_horizontal},~\ref{tab:bton_horizontal}), backbone fine-tuning methods provide the strongest accuracy-calibration trade-offs, often without requiring DCA or MDCA. On CIFAR-100 (Tab. \ref{tab:cifar_indistribution_horizontal}) without calibration regularization, LoRA achieves accuracy of 83.70\% and ECE score of 6.70, compared with the best accuracy (74.81\%) and the lowest ECE (8.91) among the prompt tuning methods. Similar advantages hold for domain and base-to-new generalization (Tabs.~\ref{tab:dg_horizontal} and~\ref{tab:bton_horizontal}). However, the benefit is not universal. LayerNorm and VeRA remain comparatively miscalibrated, showing that parameterization alone is insufficient and different factors may influence calibration.

\noindent\textbf{\emph{Role of FL optimization algorithms on calibration.}} 
We also examine whether the choice of the FL optimization algorithm has an impact on the performance of different backbone fine-tuning methods. To this end, we additionally consider specialized optimization algorithms, namely {FedProx} \cite{fedprox}, {FedDyn} \cite{feddyn} and {FedNova} \cite{fednova}. Our results are reported in Tab.~\ref{tab:fl_algo}. With FedAvg, LoRA attains the highest accuracy (83.70\%) and DoRA the lowest ECE (6.58\%). Neither result improves under FedProx, FedDyn, or FedNova. Thus, the improved calibration is primarily attributable to the choice of fine-tuning parameterization, rather than to the FL aggregation strategy, with FedAvg remaining a competitive approach. Overall, these results indicate that \textbf{well-chosen PEFT parameterizations can improve accuracy and calibration without specialized aggregation rule}.

\subsection{Does backbone fine-tuning still require post-hoc calibration?}
\label{sec:posthoc_calibration}
Having shown that well-chosen PEFT parameterizations, \textit{i.e.}, with LoRA, AdaptFormer, etc, improve calibration without
an explicit calibration objective, we next examine whether post-hoc adjustment, such as Temperature scaling ~\cite{guo2017calibration}, may provide further benefits. 

Temperature scaling divides the logits by a scalar $\tau$, typically estimated on a validation set. Fig.~\ref{fig:temp} evaluates its effect on ECE for CIFAR-10 and CIFAR-100. We analyze this aspect by comparing three representative prompt-tuning methods, namely PromptFL, FedPGP, and FedPHA, with two representative backbone fine-tuning methods, namely LoRA and DoRA. We also provide results for zero-shot CLIP (ZS-CLIP) as reference.
From Fig.~\ref{fig:temp}, we observe that moderate temperatures improve calibration, whereas overly small or large values are detrimental. ZS-CLIP is particularly sensitive: its ECE improves near $\tau=0.5$ but degrades rapidly as $\tau$ increases, indicating that moderate sharpening can be beneficial, while excessive softening induces underconfidence.
The performance of prompt tuning FL baselines varies little for $\tau\leq1$ but becomes
increasingly miscalibrated at larger temperatures; FedPGP and FedPHA degrade
most sharply, while PromptFL worsens more gradually. In contrast, LoRA and DoRA
maintain the lowest and most stable ECE across a broad temperature range, indicating reduced reliance on dataset-specific post-hoc tuning.

\begin{figure}[t]
    \centering
    \begin{subfigure}[t]{0.45\linewidth}
        \centering
        \includegraphics[width=\linewidth]{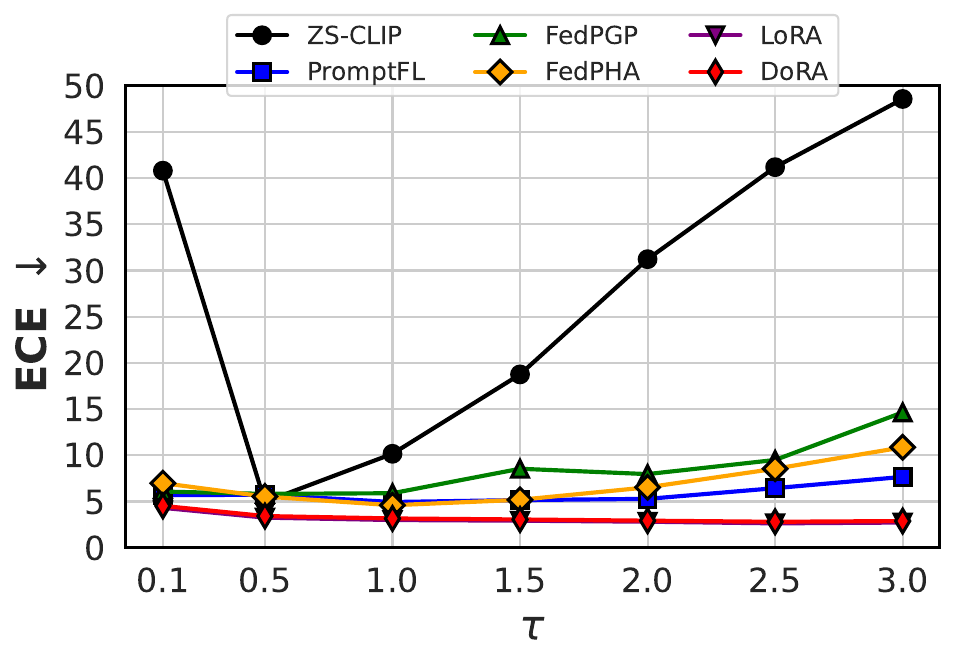}
        \caption{CIFAR-10}
        \label{fig:cifar10_tau}
    \end{subfigure}
    \hfill
    \begin{subfigure}[t]{0.45\linewidth}
        \centering
        \includegraphics[width=\linewidth]{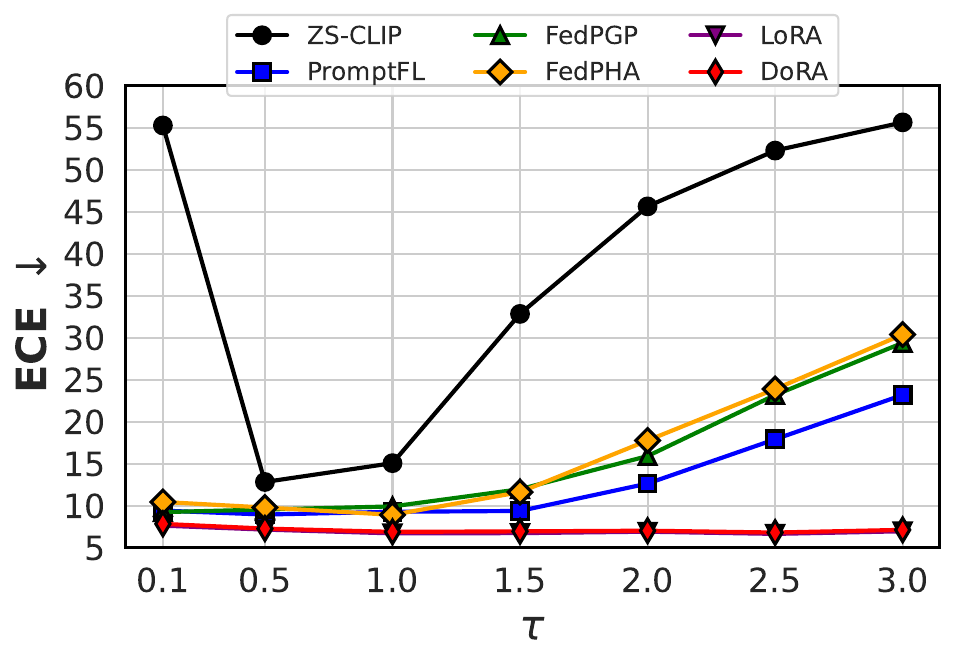}
        \caption{CIFAR-100}
        \label{fig:cifar100_tau}
    \end{subfigure}
    \vspace{-0.2cm}
    \caption{\textbf{Effect of temperature scaling ($\tau$).} We report ECE on
    (a) CIFAR-10 and (b) CIFAR-100.}
    \label{fig:temp}
    \vspace{-5pt}
\end{figure}

\subsection{Robustness under varying FL conditions}
\label{sec:fl_robustness}
Furthermore, we investigate whether the observed accuracy and calibration gains remain robust under two practical FL factors: the communication budget and client data heterogeneity. Specifically, we vary the number of communication rounds to assess the impact of the communication budget, and the Dirichlet concentration parameter to control the degree of heterogeneity in the client data distribution. As in the previous section, we compare three representative prompt-tuning methods with two backbone fine-tuning approaches, \textit{i.e.} LoRA and DoRA.

\noindent\textbf{\emph{Sensitivity to communication rounds.}} We vary the number of communication rounds from 10 to 100. As shown in Fig. \ref{fig:comround}, more rounds generally improve accuracy, while ECE stays stable or slightly improves. LoRA and DoRA consistently give the best accuracy-calibration trade-offs; even with only 10 rounds, they outperform the prompt tuning baselines evaluated after 100 rounds.

\begin{figure}[t]
    \centering
    \begin{subfigure}[t]{0.45\linewidth}
        \centering
        \includegraphics[width=\linewidth]{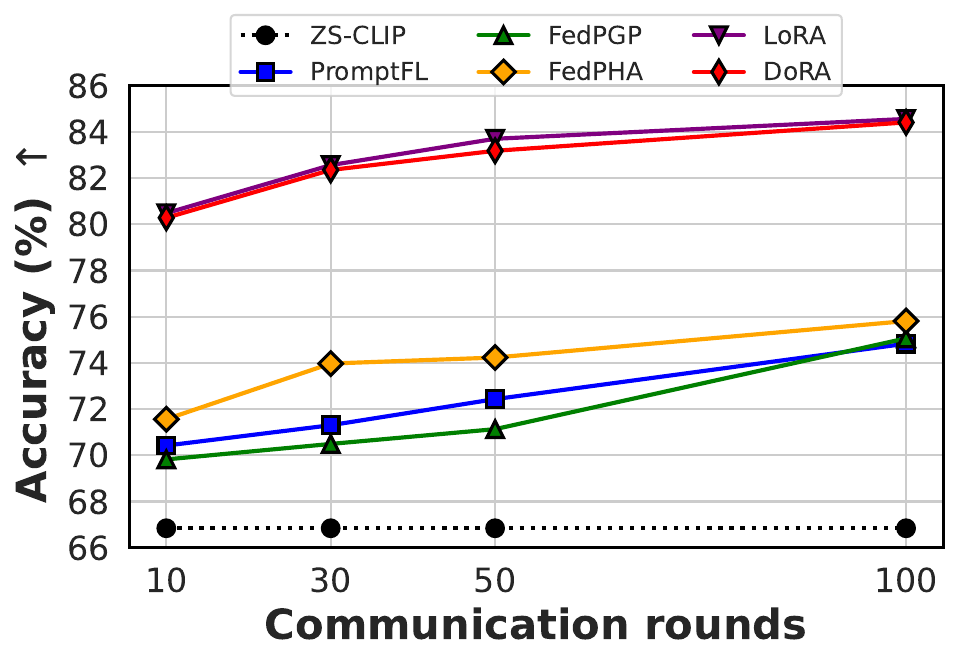}
        \caption{Accuracy}
        \label{fig:com_acc}
    \end{subfigure}
    \hfill
    \begin{subfigure}[t]{0.45\linewidth}
        \centering
        \includegraphics[width=\linewidth]{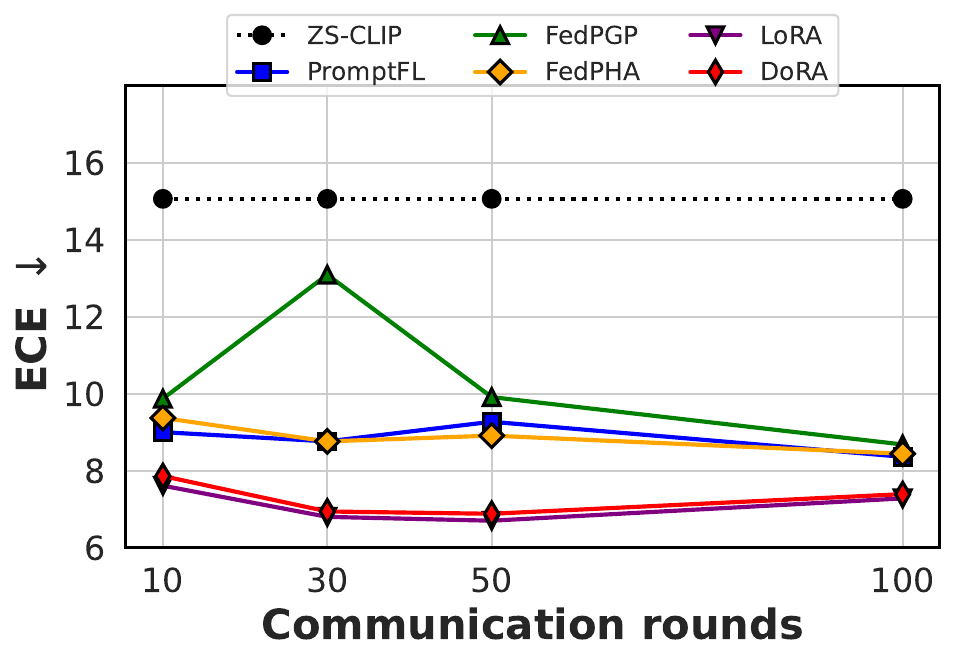}
        \caption{ECE}
        \label{fig:com_ece}
    \end{subfigure}
    \caption{\textbf{Effect of varying communication rounds}. We report (a) Accuracy and (b) ECE on CIFAR-100 dataset.}
    \label{fig:comround}
    \vspace{-10pt}
\end{figure}

\vspace{+5pt}
\noindent\textbf{\emph{Sensitivity to data distribution.}} We study the impact of varying $\beta$, the concentration parameter of the Dirichlet distribution, which controls the overlap of classes among clients. A low value of $\beta$ signifies a low overlap among clients (\textit{i.e.}, non-IID), and a high value indicates a high overlap among clients (\textit{i.e.}, IID). From Fig. \ref{fig:dirichlet}, we observe that both LoRA and DoRA remain fairly stable in terms of accuracy and ECE across varying levels of heterogeneity. Although prompt tuning methods, such as FedPGP and FedPHA, exhibit improved ECE with respect to the simple baseline ZS-CLIP, they fluctuate with different values of $\beta$.

In the \texttt{Supplementary}, we provide additional experiments covering further FL conditions, including the impact of client participation rates. We also report ablation studies on computational complexity, sensitivity to the choice of the encoder being fine-tuned, and the rank of the LoRA matrices, along with results using additional calibration metrics. 

\begin{figure}[!t]
    \centering
    \begin{subfigure}[t]{0.45\linewidth}
        \centering
        \includegraphics[width=\linewidth]{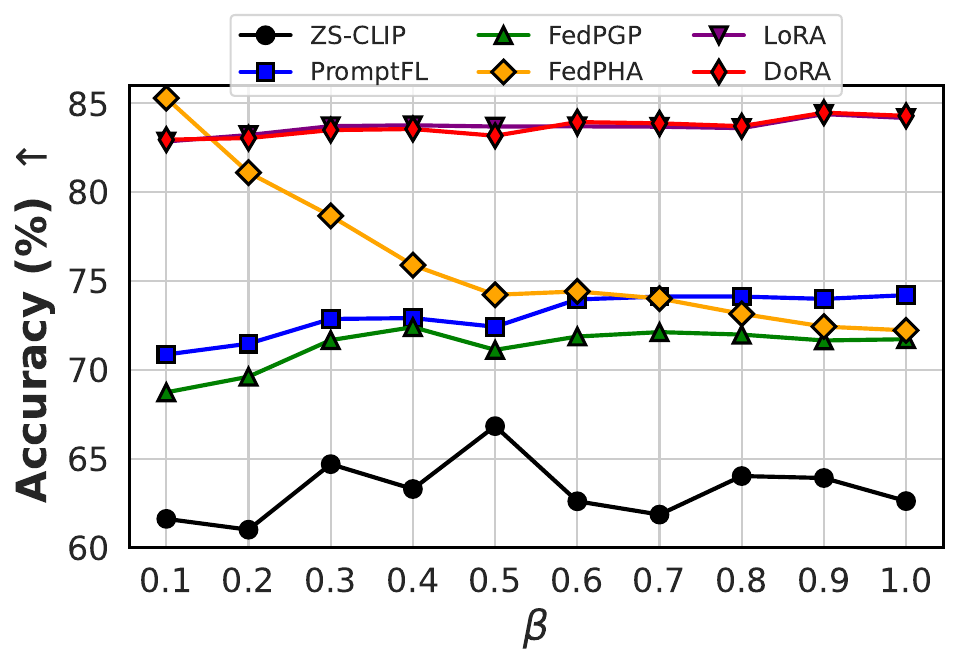}
    \end{subfigure}
    \hfill
    \begin{subfigure}[t]{0.45\linewidth}
        \centering
        \includegraphics[width=\linewidth]{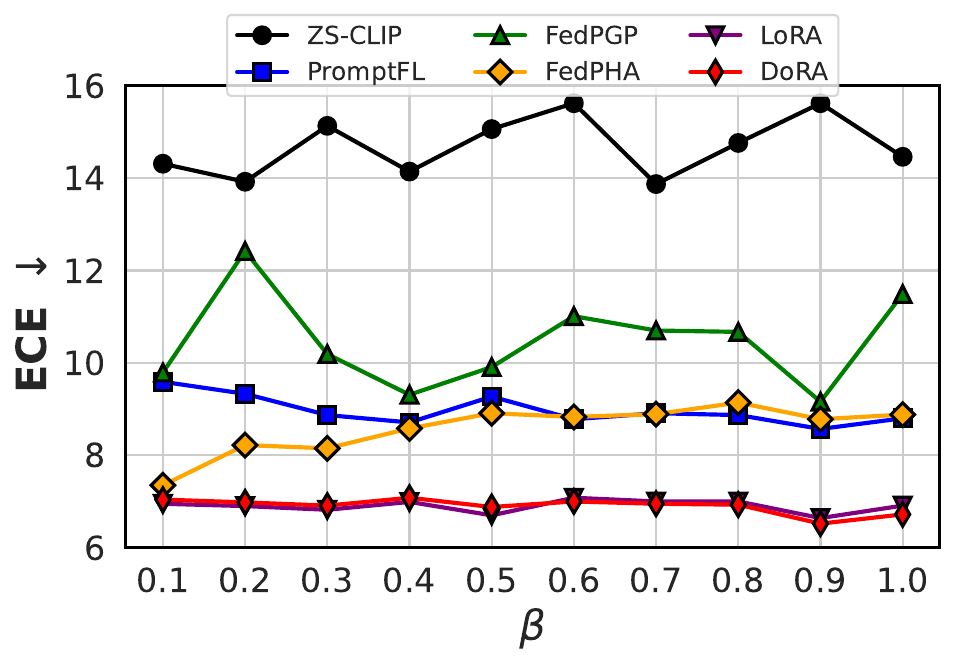}
    \end{subfigure}
    \caption{\textbf{Effect of varying $\beta$ of Dirichlet distribution}. We report (left) Accuracy and (right) ECE on CIFAR-100 dataset. }
    \label{fig:dirichlet}
    \vspace{-5pt}
\end{figure}

\section{Why are backbone fine-tuning methods better calibrated than prompt tuning?}
\label{sec:peft_prompt_calibration}

In this section we analyze how client heterogeneity and the fine-tuning parameterization
affect the difference between federated and centralized CLIP.
Let $q_n=\frac{|\mathcal D_n|}{|\mathcal D|}$ be the FedAvg sample weight in
Eq.~\eqref{eqn:fedavg} and 
$\bar g(\theta)=\sum_{n=1}^N q_n g_n(\theta)$, where
$g_n(\theta)=\nabla_\theta\mathcal L(\theta;\mathcal D_n)$ is the gradient of the local empirical loss $\mathcal{L}$ in Eq.~\eqref{eqn:ce_loss}.

Starting from the same model parameters $\theta_t$ as in Sec.~\ref{sec:prelim}, client $n$ performs $K$ local
gradient steps:
\begin{equation}
    \theta_{t,k+1}^{(n)}
    =
    \theta_{t,k}^{(n)}
    -
    \eta g_n(\theta_{t,k}^{(n)}),
    \qquad
    \theta_{t,0}^{(n)}=\theta_t,
\end{equation}
and FedAvg returns:
\(
    \theta_{t+1}^{G}
    =
    \sum_{n=1}^N q_n\theta_{t,K}^{(n)}.
\)
Similarly, the corresponding centralized model provides:
\begin{equation}
    \theta_{t,k+1}^{\mathcal{C}}
    =
    \theta_{t,k}^{\mathcal{C}}
    -
    \eta\bar g(\theta_{t,k}^{\mathcal{C}}),
    \qquad
    \theta_{t,0}^{\mathcal{C}}=\theta_t.
\end{equation}
Here, $k\in\{0,\ldots,K\}$ is the local-step
index, $\eta$ is the learning rate, and $\theta_{t,k}^{(n)}$ denotes the
parameters of client $n$ after $k$ local steps in communication round $t$. $\mathcal{G}$ and $\mathcal{C}$ denote the FedAvg global model and its centralized
counterpart, respectively: $\theta_{t+1}^{\mathcal{G}}$ is the global model returned after
the $K$ local steps, whereas $\theta_{t,k}^{\mathcal{C}}$ is the model after $k$
centralized gradient steps on $\mathcal D$.

\begin{lemma}[Federated--centralized drift]
\label{lem:fed_central_drift}
After $K$ local steps,
\begin{equation}
    \theta_{t+1}^{\mathcal{G}}-\theta_{t,K}^{\mathcal{C}}
    =
    -\eta
    \sum_{k=0}^{K-1}
    \sum_{n=1}^N q_n
    \left[
        g_n(\theta_{t,k}^{(n)})
        -
        g_n(\theta_{t,k}^{\mathcal{C}})
    \right].
    \label{eq:fed_central_drift}
\end{equation}
\end{lemma}
\textit{Proof.} See \texttt{Supplementary.}

\paragraph{The fine-tuning parameterization controls how this gap affects
the logits.}
For CLIP,
\begin{equation}
    z_c(\vect{x};\theta)
    =
    s\,
    \widehat{\mathbf v}(\vect{x};\theta)^\top
    \widehat{\mathbf t}_c(\theta).
\end{equation}
Here, $c$ is a class index, $z_c$ is the class-$c$ logit; $s$ is the CLIP
logit scale; and $\widehat{\mathbf v}$ and $\widehat{\mathbf t}_c$ are the
$\ell_2$-normalized image representation and class-$c$ text representation,
respectively. The input $\vect{x}$ follows Sec.~\ref{sec:prelim}, and
$({\cdot})^\top$ denotes transpose. Now, for a small federated-centralized parameter difference,
$\Delta^{\mathcal{G},\mathcal{C}}=\theta_{t+1}^{\mathcal{G}}-\theta_{t,K}^{\mathcal{C}}$,
\begin{equation}
    \mathbf z(\vect{x};\theta_{t+1}^{\mathcal{G}})
    -
    \mathbf z(\vect{x};\theta_{t,K}^{\mathcal{C}})
    \approx
    J_z(\vect{x};\theta_{t,K}^{\mathcal{C}})
    \Delta^{\mathcal{G},\mathcal{C}}.
    \label{eq:central_logit_gap}
\end{equation}

Here, $\mathbf z=(z_1,\ldots,z_\mathcal{C})^\top$ is the vector of class logits,
$J_z(\vect{x};\theta)=\partial\mathbf z(\vect{x};\theta)/\partial\theta$ is its
Jacobian with respect to the trainable parameters, and $\Delta^{\mathcal{G},\mathcal{C}}$ is the
federated-minus-centralized parameter gap defined above. 

In prompt tuning (PT), the trainable parameters directly modify the
class-specific text prototypes. In particular,
\begin{equation}
    \delta m_{ab}^{\mathrm{PT}}(\vect{x})
    \approx
    s\,\widehat{\mathbf v}_0(\vect{x})^\top
    (J_{\mathcal T,a}-J_{\mathcal T,b})
    \Delta\mathcal T,
\end{equation}
For class indices $a$ and $b$, let, $m_{ab}=z_a-z_b$ be their logit margin, and $\delta m_{ab}^{\mathrm{PT}}$ is its first-order federated--centralized change
under PT. So heterogeneous prompt updates can directly change class margins and
therefore confidence. Here, $\Delta\mathcal T=\mathcal T^\mathcal{G}-\mathcal T^\mathcal{C}$ is the prompt component of $\Delta^{\mathcal{G},\mathcal{C}}$.

Nevertheless, backbone fine-tuning (BFT) methods instead constrain the adaptation through residual updates,
e.g.,
\begin{equation}
    W_\ell=W_\ell^0+\Delta W_\ell(\theta).
\end{equation}
where, $\ell$ is a layer index, $W_\ell$ is the adapted layer weight, $W_\ell^0$ is its frozen pretrained value, and $\Delta W_\ell(\theta)$ is the trainable residual update parameterized by $\theta$.
However, the advantage of the updates is not guaranteed by parameter efficiency alone: it depends on whether the restricted update geometry reduces the Jacobian-weighted federated--centralized drift in Eq.~\eqref{eq:central_logit_gap}.

\begin{proposition}[Federated--centralized prediction stability]
\label{prop:centralized_stability}
Let
\(
\mathbf p^{\mathcal{G}}(\vect{x})
=
\operatorname{softmax}
\left(
    \frac{\mathbf z(\vect{x};\theta_{t+1}^{\mathcal{G}})}{\tau}
\right),
\qquad
\mathbf p^{\mathcal{C}}(\vect{x})
=
\operatorname{softmax}
\left(
    \frac{\mathbf z(\vect{x};\theta_{t,K}^{\mathcal{C}})}{\tau}
\right).
\)
where, $\mathbf p^{\mathcal{G}}$ and $\mathbf p^{\mathcal{C}}$ are the class-probability vectors of the global model and centralized comparator, respectively, and $\tau>0$ is the temperature hyperparameter. Now, if
\(
\mathcal{K}_z(\vect{x})
=
\sup_{\theta}
\|J_z(\vect{x};\theta)\|_{\mathrm{op}}
\)
along the segment between $\theta_{t+1}^{\mathcal{G}}$ and $\theta_{t,K}^{\mathcal{C}}$,
then
\begin{equation}
    \|\mathbf p^{\mathcal{G}}(\vect{x})-\mathbf p^{\mathcal{C}}(\vect{x})\|_2
    \leq
    L_\tau \mathcal{K}_z(\vect{x})
    \|\theta_{t+1}^{\mathcal{G}}-\theta_{t,K}^{\mathcal{C}}\|_2,
    \qquad \hspace{-4mm}
    L_\tau\leq\frac{1}{2\tau}.
    \label{eq:central_prediction_bound}
\end{equation}
\end{proposition}
where, $\|\cdot\|_{\mathrm{op}}$ for the operator norm induced by the Euclidean
norm $\|\cdot\|_2$; $L_\tau$ denotes the $\ell_2$-Lipschitz constant of the
temperature-scaled softmax map i.e. $\operatorname{softmax}(\cdot/\tau)$; and $\mathcal{K}_z(\vect{x})$ is the maximum logit-Jacobian sensitivity on the stated parameter segment.

\textit{Proof.} See \texttt{Supplementary.}

If the predicted classes and ECE-bin assignments remain unchanged, this
also gives
\begin{equation}
    \left|
        \operatorname{ECE}(\mathbf p^{\mathcal{G}})
        -
        \operatorname{ECE}(\mathbf p^{\mathcal{C}})
    \right|
    \leq
    \frac{L_\tau}{|\mathcal D|}
    \sum_i
    \mathcal{K}_z(\vect{x}^{(i)})
    \|\theta_{t+1}^{\mathcal{G}}-\theta_{t,K}^{\mathcal{C}}\|_2.
    \label{eq:central_ece_bound}
\end{equation}

Hence, calibration degradation in FL depends on two factors: the gap between the federated and centralized models, and the sensitivity of the fine-tuning method to this gap. Under non-IID data, local client updates can move the federated model away from the centralized solution. The Jacobian then measures how strongly this parameter gap changes the logits and predicted probabilities. Therefore, backbone fine-tuning is expected to preserve calibration better than prompt tuning when
\(
\mathcal{K}_z^{\mathrm{BFT}}(\vect{x})
\|\Delta_{\mathrm{BFT}}^{\mathcal{G},\mathcal{C}}\|_2
<
\mathcal{K}_z^{\mathrm{PT}}(\vect{x})
\|\Delta_{\mathrm{PT}}^{\mathcal{G},\mathcal{C}}\|_2,
\)
since this directly yields a tighter upper bound on the federated-centralized prediction, and consequently calibration, gap.

\section{Conclusions}
In this work, we investigated the unexplored question of how fine-tuning CLIP in a FL setting affects model calibration. Our study reveals that prompt tuning-based methods that perform reasonably in offline settings exhibit worse calibration when operating in FL. While standard in-training calibration techniques offer limited improvements, our analysis suggests that the primary driver of miscalibration is the fine-tuning of CLIP. We analyze the state-of-the-art backbone fine-tuning methods that update the intermediate representation of the encoders and show that that they naturally yield to better calibrated models in FL, substantially reducing the need for additional calibration procedures. By exploring the link between backbone fine-tuning and FL, our work opens new directions for addressing calibration of CLIP in FL setup.

{
    \small
    \bibliographystyle{ieeenat_fullname}
    \bibliography{main}

@String(CVPR  = {IEEE Conf. Comput. Vis. Pattern Recog.})

@String(ICCV  = {Int. Conf. Comput. Vis.})

@String(NeurIPS = {Adv. Neural Inform. Process. Syst.})

@String(ICML  = {Int. Conf. Mach. Learn.})

@String(ICLR  = {Int. Conf. Learn. Represent.})

@String(BMVC  = {Brit. Mach. Vis. Conf.})

@String(CVPR  = {CVPR})

@String(ICCV  = {ICCV})

@String(NeurIPS = {NeurIPS})

@String(ICML  = {ICML})

@String(ICLR  = {ICLR})

@String(BMVC  =	{BMVC})

@inproceedings{bai2024diprompt,
  title={Diprompt: Disentangled prompt tuning for multiple latent domain generalization in federated learning},
  author={Bai, Sikai and Zhang, Jie and Guo, Song and Li, Shuaicheng and Guo, Jingcai and Hou, Jun and Han, Tao and Lu, Xiaocheng},
  booktitle={Proceedings of the IEEE/CVF Conference on Computer Vision and Pattern Recognition},
  pages={27284--27293},
  year={2024}
}

@article{wang2024open,
  title={Open-vocabulary calibration for fine-tuned CLIP},
  author={Wang, Shuoyuan and Wang, Jindong and Wang, Guoqing and Zhang, Bob and Zhou, Kaiyang and Wei, Hongxin},
  journal={International Conference on Machine Learning (ICML)},
  year={2024}
}

@article{minderer2021revisiting,
  title={Revisiting the calibration of modern neural networks},
  author={Minderer, Matthias and Djolonga, Josip and Romijnders, Rob and Hubis, Frances and Zhai, Xiaohua and Houlsby, Neil and Tran, Dustin and Lucic, Mario},
  journal={Advances in neural information processing systems},
  volume={34},
  pages={15682--15694},
  year={2021}
}

@inproceedings{bohdal2025efficient,
  title={Efficient compositional multi-tasking for on-device large language models},
  author={Bohdal, Ondrej and Ozay, Mete and Moon, Jijoong and Lee, Kyenghun and Ko, Hyeonmok and Michieli, Umberto},
  booktitle={Proceedings of the 2025 Conference on Empirical Methods in Natural Language Processing},
  pages={28129--28153},
  year={2025}
}

@inproceedings{hsu2019measuring,
  title={Measuring the effects of non-identical data distribution for federated visual classification},
  author={Hsu, Tzu-Ming Harry and Qi, Hang and Brown, Matthew},
  booktitle={NeurIPS Workshop on Federated Learning},
  year={2019}
}

@inproceedings{khan2025confidence,
  title={Confidence-calibrated covariate shift correction for few-shot classification in Vision-Language Models},
  author={Khan, Behraj and Qureshi, Rizwan and Durrani, Nouman Muhammad and Syed, Tahir Qasim},
  booktitle={Proceedings of the Computer Vision and Pattern Recognition Conference},
  pages={6511--6523},
  year={2025}
}

@article{coop,
  title={Learning to prompt for vision-language models},
  author={Zhou, Kaiyang and Yang, Jingkang and Loy, Chen Change and Liu, Ziwei},
  journal={International Journal of Computer Vision},
  volume={130},
  number={9},
  pages={2337--2348},
  year={2022},
  publisher={Springer}
}

@inproceedings{lin2017focal,
  title={Focal loss for dense object detection},
  author={Lin, Tsung-Yi and Goyal, Priya and Girshick, Ross and He, Kaiming and Doll{\'a}r, Piotr},
  booktitle={Proceedings of the IEEE international conference on computer vision},
  pages={2980--2988},
  year={2017}
}

@article{peng2024fedcal,
  title={Fedcal: Achieving local and global calibration in federated learning via aggregated parameterized scaler},
  author={Peng, Hongyi and Yu, Han and Tang, Xiaoli and Li, Xiaoxiao},
  journal={International Conference on Machine Learning (ICML)},
  year={2024}
}

@inproceedings{hebbalaguppe2022stitch,
  title={A stitch in time saves nine: A train-time regularizing loss for improved neural network calibration},
  author={Hebbalaguppe, Ramya and Prakash, Jatin and Madan, Neelabh and Arora, Chetan},
  booktitle={Proceedings of the IEEE/CVF Conference on Computer Vision and Pattern Recognition},
  pages={16081--16090},
  year={2022}
}

@article{muller2019does,
  title={When does label smoothing help?},
  author={M{\"u}ller, Rafael and Kornblith, Simon and Hinton, Geoffrey E},
  journal={Advances in neural information processing systems},
  volume={32},
  year={2019}
}

@inproceedings{kull2017beta,
  title={Beta calibration: a well-founded and easily implemented improvement on logistic calibration for binary classifiers},
  author={Kull, Meelis and Silva Filho, Telmo and Flach, Peter},
  booktitle={Artificial intelligence and statistics},
  pages={623--631},
  year={2017},
  organization={PMLR}
}

@inproceedings{nixon2019measuring,
  title={Measuring calibration in deep learning.},
  author={Nixon, Jeremy and Dusenberry, Michael W and Zhang, Linchuan and Jerfel, Ghassen and Tran, Dustin},
  booktitle={CVPR workshops},
  volume={2},
  number={7},
  year={2019}
}

@inproceedings{shamsian2021personalized,
  title={Personalized federated learning using hypernetworks},
  author={Shamsian, Aviv and Navon, Aviv and Fetaya, Ethan and Chechik, Gal},
  booktitle={International conference on machine learning},
  pages={9489--9502},
  year={2021},
  organization={PMLR}
}

@inproceedings{cao2023knowledge,
  title={Knowledge-aware federated active learning with non-iid data},
  author={Cao, Yu-Tong and Shi, Ye and Yu, Baosheng and Wang, Jingya and Tao, Dacheng},
  booktitle={Proceedings of the IEEE/CVF International Conference on Computer Vision},
  pages={22279--22289},
  year={2023}
}

@inproceedings{guo2017calibration,
  title={On calibration of modern neural networks},
  author={Guo, Chuan and Pleiss, Geoff and Sun, Yu and Weinberger, Kilian Q},
  booktitle={International conference on machine learning},
  pages={1321--1330},
  year={2017},
  organization={PMLR}
}

@article{pavlovic2025understanding,
  title={Understanding Model Calibration--A gentle introduction and visual exploration of calibration and the expected calibration error (ECE)},
  author={Pavlovic, Maja},
  journal={International Conference on Learning Representations Blogposts},
  year={2025}
}

@article{yang2025federated,
  title={Federated Low-Rank Adaptation for Foundation Models: A Survey},
  author={Yang, Yiyuan and Long, Guodong and Lu, Qinghua and Zhu, Liming and Jiang, Jing and Zhang, Chengqi},
  journal={arXiv preprint arXiv:2505.13502},
  year={2025}
}

@inproceedings{pacs,
  title={Deeper, broader and artier domain generalization},
  author={Li, Da and Yang, Yongxin and Song, Yi-Zhe and Hospedales, Timothy M},
  booktitle={2017 IEEE International Conference on Computer Vision (ICCV)},
  pages={5543--5551},
  year={2017},
  organization={Ieee}
}

@inproceedings{li2017deeper,
  title={Deeper, broader and artier domain generalization},
  author={Li, Da and Yang, Yongxin and Song, Yi-Zhe and Hospedales, Timothy M},
  booktitle={Proceedings of the IEEE international conference on computer vision},
  pages={5542--5550},
  year={2017}
}

@article{fine-tune,
  title={Fine-tuning can distort pretrained features and underperform out-of-distribution},
  author={Kumar, Ananya and Raghunathan, Aditi and Jones, Robbie and Ma, Tengyu and Liang, Percy},
  journal={arXiv preprint arXiv:2202.10054},
  year={2022}
}

@inproceedings{food,
  title={Food-101 - Mining Discriminative Components with Random Forests},
  author={Lukas Bossard and Matthieu Guillaumin and Luc Van Gool},
  booktitle={European Conference on Computer Vision},
  year={2014}
}

@inproceedings{vpt,
  title={Visual prompt tuning},
  author={Jia, Menglin and Tang, Luming and Chen, Bor-Chun and Cardie, Claire and Belongie, Serge and Hariharan, Bharath and Lim, Ser-Nam},
  booktitle={European Conference on Computer Vision},
  pages={709--727},
  year={2022},
  organization={Springer}
}

@article{krizhevsky2012imagenet,
  title={Imagenet classification with deep convolutional neural networks},
  author={Krizhevsky, Alex and Sutskever, Ilya and Hinton, Geoffrey E},
  journal={Advances in neural information processing systems},
  volume={25},
  year={2012}
}

@article{labelme,
  title={LabelMe: a database and web-based tool for image annotation},
  author={Russell, Bryan C and Torralba, Antonio and Murphy, Kevin P and Freeman, William T},
  journal={International journal of computer vision},
  volume={77},
  number={1},
  pages={157--173},
  year={2008},
  publisher={Springer}
}

@inproceedings{radford2019language,
  title={Learning transferable visual models from natural language supervision},
  author={Radford, Alec and Kim, Jong Wook and Hallacy, Chris and Ramesh, Aditya and Goh, Gabriel and Agarwal, Sandhini and Sastry, Girish and Askell, Amanda and Mishkin, Pamela and Clark, Jack and others},
  booktitle={International Conference on Machine Learning},
  pages={8748--8763},
  year={2021},
  organization={PMLR}
}

@inproceedings{officehome,
  title={Deep hashing network for unsupervised domain adaptation},
  author={Venkateswara, Hemanth and Eusebio, Jose and Chakraborty, Shayok and Panchanathan, Sethuraman},
  booktitle={Proceedings of the IEEE Conference on Computer Vision and Pattern Recognition},
  pages={5018--5027},
  year={2017}
}

@article{csurka2017domain,
  title={Domain adaptation for visual applications: A comprehensive survey},
  author={Csurka, Gabriela},
  journal={arXiv preprint arXiv:1702.05374},
  year={2017}
}

@InProceedings{maple,
    author    = {Khattak, Muhammad Uzair and Rasheed, Hanoona and Maaz, Muhammad and Khan, Salman and Khan, Fahad Shahbaz},
    title     = {MaPLe: Multi-Modal Prompt Learning},
    booktitle = {Proceedings of the IEEE/CVF Conference on Computer Vision and Pattern Recognition (CVPR)},
    month     = {June},
    year      = {2023},
    pages     = {19113-19122}
}

@inproceedings{dtd,
  title={Describing textures in the wild},
  author={Cimpoi, Mircea and Maji, Subhransu and Kokkinos, Iasonas and Mohamed, Sammy and Vedaldi, Andrea},
  booktitle={Proceedings of the IEEE Conference on Computer Vision and Pattern Recognition},
  pages={3606--3613},
  year={2014}
}

@InProceedings{vlcs,
author = {Fang, Chen and Xu, Ye and Rockmore, Daniel N.},
title = {Unbiased Metric Learning: On the Utilization of Multiple Datasets and Web Images for Softening Bias},
booktitle = {Proceedings of the IEEE International Conference on Computer Vision (ICCV)},
month = {December},
year = {2013}
}

@inproceedings{caltech,
  title={Learning generative visual models from few training examples: An incremental bayesian approach tested on 101 object categories},
  author={Fei-Fei, Li and Fergus, Rob and Perona, Pietro},
  booktitle={Conference on Computer Vision and Pattern Recognition Workshop},
  pages={178--178},
  year={2004},
  organization={IEEE}
}

@inproceedings{lora,
  title={Lora: Low-rank adaptation of large language models.},
  author={Hu, Edward J and Shen, Yelong and Wallis, Phillip and Allen-Zhu, Zeyuan and Li, Yuanzhi and Wang, Shean and Wang, Lu and Chen, Weizhu and others},
  booktitle={International Conference on Learning Representations},
  volume={1},
  number={2},
  pages={3},
  year={2022}
}

@inproceedings{he2016deep,
  title={Deep residual learning for image recognition},
  author={He, Kaiming and Zhang, Xiangyu and Ren, Shaoqing and Sun, Jian},
  booktitle={Proceedings of the IEEE conference on computer vision and pattern recognition},
  pages={770--778},
  year={2016}
}

@inproceedings{geng2024survey,
  title={A survey of confidence estimation and calibration in large language models},
  author={Geng, Jiahui and Cai, Fengyu and Wang, Yuxia and Koeppl, Heinz and Nakov, Preslav and Gurevych, Iryna},
  booktitle={Proceedings of the 2024 Conference of the North American Chapter of the Association for Computational Linguistics: Human Language Technologies (Volume 1: Long Papers)},
  pages={6577--6595},
  year={2024}
}

@article{wang2023calibration,
  title={Calibration in deep learning: A survey of the state-of-the-art},
  author={Wang, Cheng},
  journal={arXiv preprint arXiv:2308.01222},
  year={2023}
}

@inproceedings{kgcoop,
  title={Visual-language prompt tuning with knowledge-guided context optimization},
  author={Yao, Hantao and Zhang, Rui and Xu, Changsheng},
  booktitle={Proceedings of the IEEE/CVF Conference on Computer Vision and Pattern Recognition},
  pages={6757--6767},
  year={2023}
}

@InProceedings{oxfordpets,
  author       = "Omkar M. Parkhi and Andrea Vedaldi and Andrew Zisserman and C. V. Jawahar",
  title        = "Cats and Dogs",
  booktitle    = "IEEE Conference on Computer Vision and Pattern Recognition",
  year         = "2012",
}

@INPROCEEDINGS{oxfordflowers,

  author={Nilsback, Maria-Elena and Zisserman, Andrew},

  booktitle={2008 Sixth Indian Conference on Computer Vision, Graphics \& Image Processing}, 

  title={Automated Flower Classification over a Large Number of Classes}, 

  year={2008},

  volume={},

  number={},

  pages={722-729},

  doi={10.1109/ICVGIP.2008.47}}

@inproceedings{food101,
  title={Food-101 - Mining Discriminative Components with Random Forests},
  author={Lukas Bossard and Matthieu Guillaumin and Luc Van Gool},
  booktitle={European Conference on Computer Vision},
  year={2014}
}

@article{promptfl,
  title={Promptfl: Let federated participants cooperatively learn prompts instead of models-federated learning in age of foundation model},
  author={Guo, Tao and Guo, Song and Wang, Junxiao and Tang, Xueyang and Xu, Wenchao},
  journal={IEEE Transactions on Mobile Computing},
  year={2023},
  publisher={IEEE}
}

@inproceedings{sun2024towards,
  title={Towards multi-modal transformers in federated learning},
  author={Sun, Guangyu and Mendieta, Matias and Dutta, Aritra and Li, Xin and Chen, Chen},
  booktitle={European Conference on Computer Vision},
  pages={229--246},
  year={2024},
  organization={Springer}
}

@inproceedings{qiu2024federated,
  title={Federated text-driven prompt generation for vision-language models},
  author={Qiu, Chen and Li, Xingyu and Mummadi, Chaithanya Kumar and Ganesh, Madan Ravi and Li, Zhenzhen and Peng, Lu and Lin, Wan-Yi},
  booktitle={The Twelfth International Conference on Learning Representations},
  year={2024}
}

@inproceedings{fedclip,
  title={FedCLIP: Fast Generalization and Personalization for CLIP in Federated Learning},
  author={Lu, Wang and Xixu, HU and Wang, Jindong and Xie, Xing},
  booktitle={ICLR Workshop on Trustworthy and Reliable Large-Scale Machine Learning Models},
  year={2023}
}

@inproceedings{fedotp,
  title={Global and local prompts cooperation via optimal transport for federated learning},
  author={Li, Hongxia and Huang, Wei and Wang, Jingya and Shi, Ye},
  booktitle={Proceedings of the IEEE/CVF Conference on Computer Vision and Pattern Recognition},
  pages={12151--12161},
  year={2024}
}

@article{dor,
  title={Understanding and mitigating miscalibration in prompt tuning for vision-language models},
  author={Wang, Shuoyuan and Li, Yixuan and Wei, Hongxin},
  journal={International Conference on Machine Learning},
  year={2025}
}

@inproceedings{fedavg,
  title={Communication-efficient learning of deep networks from decentralized data},
  author={McMahan, Brendan and Moore, Eider and Ramage, Daniel and Hampson, Seth and y Arcas, Blaise Aguera},
  booktitle={Artificial intelligence and statistics},
  pages={1273--1282},
  year={2017},
  organization={PMLR}
}

@article{fedprox,
  title={Federated optimization in heterogeneous networks},
  author={Li, Tian and Sahu, Anit Kumar and Zaheer, Manzil and Sanjabi, Maziar and Talwalkar, Ameet and Smith, Virginia},
  journal={Proceedings of Machine learning and systems},
  volume={2},
  pages={429--450},
  year={2020}
}

@article{feddyn,
  title={Federated learning based on dynamic regularization},
  author={Acar, Durmus Alp Emre and Zhao, Yue and Navarro, Ramon Matas and Mattina, Matthew and Whatmough, Paul N and Saligrama, Venkatesh},
  journal={International Conference on Learning Representations},
  year={2021}
}

@article{fednova,
  title={Tackling the objective inconsistency problem in heterogeneous federated optimization},
  author={Wang, Jianyu and Liu, Qinghua and Liang, Hao and Joshi, Gauri and Poor, H Vincent},
  journal={Advances in neural information processing systems},
  volume={33},
  pages={7611--7623},
  year={2020}
}

@article{yang2024generalized,
  title={Generalized out-of-distribution detection: A survey},
  author={Yang, Jingkang and Zhou, Kaiyang and Li, Yixuan and Liu, Ziwei},
  journal={International Journal of Computer Vision},
  volume={132},
  number={12},
  pages={5635--5662},
  year={2024},
  publisher={Springer}
}

@inproceedings{fedmvp,
  title={FedMVP: Federated Multimodal Visual Prompt Tuning for Vision-Language Models},
  author={Singha, Mainak and Roy, Subhankar and Mehrotra, Sarthak and Jha, Ankit and Abdar, Moloud and Banerjee, Biplab and Ricci, Elisa},
  booktitle={Proceedings of the IEEE/CVF International Conference on Computer Vision},
  pages={17869--17878},
  year={2025}
}

@article{nucfl,
  title={Unlocking the potential of model calibration in federated learning},
  author={Chu, Yun-Wei and Han, Dong-Jun and Hosseinalipour, Seyyedali and Brinton, Christopher},
  journal={International Conference on Learning Representation (ICLR)},
  year={2025}
}

@article{dca,
  title={Improved trainable calibration method for neural networks on medical imaging classification},
  author={Liang, Gongbo and Zhang, Yu and Wang, Xiaoqin and Jacobs, Nathan},
  journal={British Machine Vision Conference (BMVC)},
  year={2020}
}

@inproceedings{fedpha,
  title={FedPHA: Federated Prompt Learning for Heterogeneous Client Adaptation},
  author={Fang, Chengying and Huang, Wenke and Wan, Guancheng and Yang, Yihao and Ye, Mang},
  booktitle={Forty-second International Conference on Machine Learning}
}

@article{fedpgp,
  title={Harmonizing generalization and personalization in federated prompt learning},
  author={Cui, Tianyu and Li, Hongxia and Wang, Jingya and Shi, Ye},
  journal={International Conference on Machine Learning},
  year={2024}
}

@article{cifar,
  title={Learning multiple layers of features from tiny images},
  author={Krizhevsky, Alex and Hinton, Geoffrey and others},
  year={2009},
  publisher={Toronto, ON, Canada}
}

@article{sgd,
  title={A stochastic approximation method},
  author={Robbins, Herbert and Monro, Sutton},
  journal={The annals of mathematical statistics},
  pages={400--407},
  year={1951},
  publisher={JSTOR}
}

@article{layernorm,
  title={How to adapt your large-scale vision-and-language model},
  author={Kim, Konwoo and Laskin, Michael and Mordatch, Igor and Pathak, Deepak},
  year={2021}
}

@book{bayesian,
  title={Bayesian data analysis},
  author={Gelman, Andrew and Carlin, John B and Stern, Hal S and Rubin, Donald B},
  year={1995},
  publisher={Chapman and Hall/CRC}
}

@article{adaptformer,
  title={Adaptformer: Adapting vision transformers for scalable visual recognition},
  author={Chen, Shoufa and Ge, Chongjian and Tong, Zhan and Wang, Jiangliu and Song, Yibing and Wang, Jue and Luo, Ping},
  journal={Advances in Neural Information Processing Systems},
  volume={35},
  pages={16664--16678},
  year={2022}
}

@inproceedings{dora,
  title={Dora: Weight-decomposed low-rank adaptation},
  author={Liu, Shih-Yang and Wang, Chien-Yi and Yin, Hongxu and Molchanov, Pavlo and Wang, Yu-Chiang Frank and Cheng, Kwang-Ting and Chen, Min-Hung},
  booktitle={Forty-first International Conference on Machine Learning},
  year={2024}
}

@article{vera,
  title={Vera: Vector-based random matrix adaptation},
  author={Kopiczko, Dawid J and Blankevoort, Tijmen and Asano, Yuki M},
  journal={arXiv preprint arXiv:2310.11454},
  year={2023}
}
}
\newpage
\appendix

\section*{Table of Supplementary Contents}
\small{
\startcontents[chapters]
\printcontents[chapters]{}{1}{}
}

\section{Experimental setup}
\label{experimental_setup}

\subsection{Benchmarks}
In our experiments, we evaluate our proposed method across three different federated learning (FL) settings using a diverse collection of benchmark datasets, as summarized in Table \ref{tab:datasets}. For the \textit{in-distribution} setting, we use CIFAR-10 and CIFAR-100, which are standard image classification datasets with 10 and 100 classes, respectively. Each dataset contains 50,000 training images and 10,000 test images, providing a balanced evaluation scenario where the training and test distributions are identical.

For the \textit{base-to-new generalization} and \textit{domain generalization} settings, we consider more challenging datasets that assess the model's ability to transfer knowledge to unseen classes or between multiple domains. In the \textit{base-to-new} setting, we choose Caltech101, Flowers102, OxfordPets, Food101, and DTD datasets in a few-shot training setup. In the \textit{domain generalization (DG)} setting, we use traditional DG benchmarks having four domains each i.e. PACS (art painting, cartoon, photo and sketch), OfficeHome (art, clipart, product and realworld), and VLCS (caltech, labelme, pascal-voc and sun), providing distribution shifts.

\begin{table}[]
    \centering
    \caption{\textbf{Statistics of the benchmarks}. The table shows the statistics of the benchmarks used for empirical evaluation in three federated learning settings.} 
    \scalebox{0.75}{
            \begin{tabular}{lc|cccc}
            \toprule
        
         \textbf{Benchmarks} & \textbf{Domain} & {Classes} & {Train} & {Test}\\ 
         \midrule
         \rowcolor{gray!30}
         \multicolumn{5}{c}{\textit{In-distribution} setting} \\
         CIFAR-10 \cite{cifar} & - &10 &50,000 &10,000 \\
         CIFAR-100 \cite{cifar} & - &100 &50,000 &10,000 \\
         \midrule
         \rowcolor{gray!30}
         \multicolumn{5}{c}{\textit{Base-to-new generalization} setting} \\
         Caltech101 \cite{caltech} & - &101 &4,128 &2,465 \\
         Flowers102 \cite{oxfordflowers} & - &102 &4,093 &2,463 \\
         OxfordPets \cite{oxfordpets} & - &37 &2,944 &3,369 \\
         Food101 \cite{food101} & - &101 &50,500 &30,300 \\
         DTD \cite{dtd} & - & 47 &2,820 &1,692\\
   
         \midrule
         \rowcolor{gray!30}
         \multicolumn{5}{c}{\textit{Domain generalization} setting} \\
         \multirow{4}{*}{PACS \cite{li2017deeper}} &Art Painting &\multirow{4}{*}{7} &1,024	&614\\
         &Cartoon & &1,171	&704\\
         &Photo & &835	&502\\
         &Sketch & &1,964	&1,179\\
         \hline
         
         \multirow{4}{*}{OfficeHome \cite{officehome}} &Art &\multirow{4}{*}{65} &1,214	&728 \\
         &Clipart & &2,191	&1,298	\\
         &Product & &2,226	&1,324	\\
         &RealWorld & &2,180	&1,304	\\
         \hline
         
         \multirow{4}{*}{VLCS \cite{vlcs}} &CALTECH &\multirow{4}{*}{5} &891	&424	\\
         &LABELME & &1,672	&797	\\
         &PASCAL-VOC & &2,127	&1,013 \\
         &SUN & &2,067	&985	\\

        \bottomrule
        \end{tabular}}
    \label{tab:datasets}
\end{table}

\begin{figure*}[!h]
    \centering
    \includegraphics[width=\linewidth]{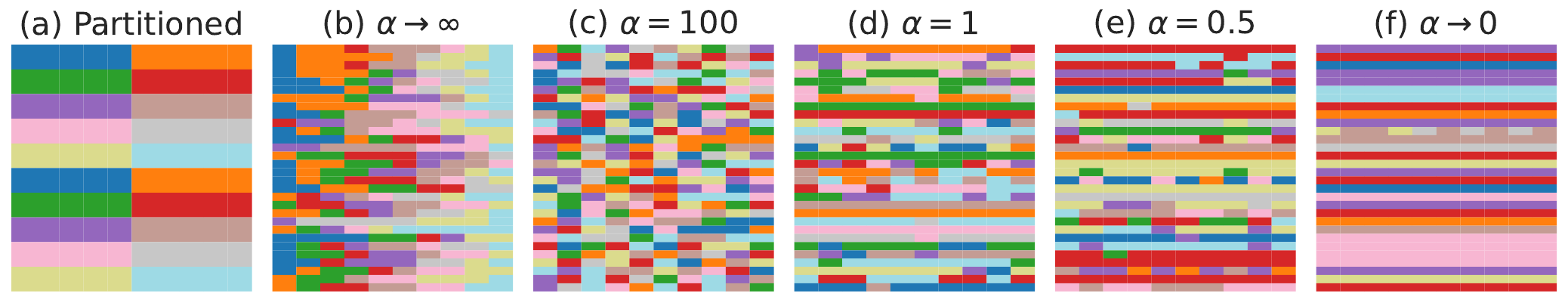}
    \vspace{-0.3cm}
    \caption{\textbf{Federated data partitioning under different Dirichlet concentration parameter $\alpha$.} In each of the subplots, each row represents a client, and each cell corresponds to a data sample whose color indicates its class label. The sub figures are represented as, (a) Sort and partition setting with two fixed classes per client, (b) to (f) Non-IID splits with Dirichlet distribution, having $\alpha \!\to\! \infty$, $\alpha = 100, 1, 0.5$, and $\alpha \!\to\! 0$. Larger $\alpha$ produces nearly IID client distributions, whereas smaller $\alpha$ concentrates samples from few classes only within each client with increasingly skewed and heterogeneous partitions.}
    \label{fig:class_dist}
\end{figure*}

\subsection{Federated learning settings}
\label{sec:flsetup}
We describe in detail the three personalized Federated Learning:
\textit{in-distribution (ID)}, \textit{domain generalization (DG)} and \textit{base-to-new generalization}.

[-] \textbf{In-distribution \& Domain generalization.} For both settings, we distribute the classes across clients in an overlapping non-independent and identically Distributed (non-IID) manner, as similar to \cite{fedotp}. To simulate controllable non-IID data heterogeneity across clients, we adopt the standard Dirichlet-based partitioning strategy \cite{fedotp, cao2023knowledge, shamsian2021personalized}. Given a dataset with $C$ classes and $N$ clients, each client's class proportion vector, $\mathbf{q}_n = (q_{n1}, \dots, q_{nK})$ is sampled from a symmetric Dirichlet distribution \cite{bayesian} parameterized by concentration $\alpha$:
\begin{equation}
    \mathbf{q}_n \sim \mathrm{Dirichlet}(\alpha \mathbf{p}),
\label{eq:dirichlet1}
\end{equation}
where $\mathbf{p}$ is the global class prior (assumed uniform in our experiments). Once the class proportions are obtained, the actual number of samples assigned to client $n$ is drawn as,
\begin{equation}
    \mathbf{c}_n \sim \mathrm{Multinomial}(|\mathcal{D}_n|, \mathbf{q}_n),
\label{eq:dirichlet2}
\end{equation}
where $|\mathcal{D}_n|$ denotes the local dataset size. The concentration parameter $\alpha$ directly controls the heterogeneity of the federated setting i.e. large $\alpha$ values (e.g., $\alpha \!\rightarrow\! \infty$) yield nearly IID client distributions, while small $\alpha$ values (e.g., $\alpha \!\rightarrow\! 0$) produce extremely skewed, label-concentrated clients. Figure~\ref{fig:class_dist} visualizes this progression, showing how the distribution transitions 
from highly mixed to strongly partitioned as $\alpha$ decreases. We set $\alpha = 0.5$ for both federated settings to maintain a balanced, overlapping non-IID distribution.
For in-distribution, we choose 100 clients over a 10\% participation rate, whereas two clients per domain with 100\% participation rate in domain generalization setting.

[-] \textbf{Base-to-New generalization.} Unlike the Dirichlet distribution, we adopt a non-overlapping non-IID setting, where each client is assigned a distinct subset of classes, following the similar set up of \cite{fedpgp}. Half of the total classes are used for training, while the remaining classes are reserved for evaluation. We configure 10 clients per dataset, with a 100\% participation rate for all clients.

\subsection{Evaluation protocol and metrics} 
\noindent\textbf{Evaluation protocol}. We evaluate the methods in the three aforementioned FL settings in a personalized FL setup \cite{fedotp, fedpha, fedpgp}, where we compute the metrics on the test set in each client and then take an average over all the clients. To be noted that client participation in training may vary according to the participation rate, but inference is performed on all clients regardless of their participation.

\noindent\textbf{Metrics}. In our FL experiments, we use the accuracy metric for model performance and ECE, MCE, and ACE metrics for model calibration.

[-] \textbf{Accuracy.} It measures the proportion of predictions for which the model's predicted label matches the ground-truth label, providing a standard indicator of overall classification performance.

[-] \textbf{Expected Calibration Error (ECE).} ECE  \cite{guo2017calibration} measures the average discrepancy between a model’s predicted confidence and accuracy. The prediction confidences are partitioned into $G$ static bins, and for each bin $b_g$, the absolute difference between the average accuracy $\text{acc}(b_g)$ and the mean predicted confidence $\text{conf}(b_g)$ is computed. These bin-wise errors are then weighted by the proportion of samples in each bin, yielding a global estimate of how well the model's confidence reflects the true likelihood of correctness. Lower ECE values indicate better-calibrated predictions. The ECE metric can be defined as:
\begin{equation}
    \text{ECE} = \sum^{G}_{g=1} \frac{\mid b_g \mid}{|\mathcal{D}|} \mid \text{acc}(b_g) - \text{conf}(b_g) \mid,
\label{eq:ece}
\end{equation}
where $|\mathcal{D}|$ and $|b_g|$ is the total number of samples in the dataset and $g$-th bin, respectively.

[-] \textbf{Maximum Calibration Error (MCE).} MCE  \cite{guo2017calibration} quantifies the worst-case miscalibration across all bins. Rather than averaging, MCE takes the maximum absolute gap between accuracy and confidence over the $G$ static bins. This highlights extreme local over- or under-confidence that may be masked by aggregate metrics such as ECE. A lower MCE value indicates that the model avoids severe calibration failures in any confidence region. The MCE metric can be written as:

\begin{equation}
    \text{MCE} = \max_{g \in \{1,\dots,G\}} \mid \text{acc}(b_g) - \text{conf}(b_g) \mid.
\label{eq:mce}
\end{equation}

[-] \textbf{Adaptive Calibration Error (ACE).} 
ACE \cite{nixon2019measuring} assesses calibration by computing the discrepancy between predicted confidence and accuracy using adaptive binning. Unlike ECE which divides the predictions into $G$ static bins over fixed confidence intervals, ACE constructs $H$ adaptive bins by sorting the predicted probabilities and dividing them into bins having an equal number of samples, to provide a more realistic calculation of calibration across the entire confidence distribution. ACE evaluates calibration across all class probabilities rather than considering only the maximum predicted confidence. The ACE metric can be defined as:

\begin{equation}
    \text{ACE} = \frac{1}{C}\frac{1}{H} \sum^{C}_{c=1}\sum^{H}_{h=1} \mid \text{acc}(b_{h,c}) - \text{conf}(b_{h,c}) \mid,
\label{eq:ace}
\end{equation}

where $ \text{acc}(b_{h,c})$ and $\text{conf}(b_{h,c})$ are respectively the average accuracy and  predicted probability for class $c$ among the samples belonging to bin $h$. %

\subsection{Calibration baselines}
In our experimental settings, we consider two standalone calibration methods: DCA \cite{dca} and MDCA \cite{hebbalaguppe2022stitch} as additional plug-ins in the federated methods.

[-] \textbf{Difference-between-Confidence-Accuracy (DCA).}
DCA \cite{dca} provides a minibatch-level calibration objective by penalizing the discrepancy between the model's average predicted confidence score corresponding to the true class and its average accuracy for all the samples in a minibatch. For each prediction corresponding to the $i-$th sample, the method computes the correctness indicator $c_i$ (which equals $1$ if the prediction is correct and $0$ otherwise) and the predicted confidence of the true label $s_i = p(\hat{y}_i)$. By minimizing this deviation, DCA encourages the model's overall confidence level to align with its average true accuracy, thereby improving global calibration behavior. For a minibatch of size $m$, the loss is computed as:

\begin{equation}
    \mathcal{L}_{\mathrm{dca}}
    = 
     \Big\lvert \frac{1}{m} \sum_{i=1}^m 
        c_i 
        - 
        \frac{1}{m} \sum_{i=1}^m s_i
    \Big\lvert.
    \label{eq:dca}
\end{equation}

Eq. \ref{eq:dca} computes an auxiliary calibration loss by taking, over $m$ samples in a minibatch, that penalizes the mismatch between observed mean correctness and mean confidence, thereby guiding the model to better align its confidence estimates with accuracy, without a strict constraint on each example. Note that during backpropagation, the gradients are backpropagated only through the confidence term $s_i$.

[-] \textbf{Multi-class Difference-between-Confidence-Accuracy (MDCA).} MDCA \cite{hebbalaguppe2022stitch} generalizes DCA to the multi-class setting by evaluating calibration across all class probabilities, rather than only the predicted class. For each class $j$, the method computes the gap between the average one-hot ground-truth indicator $c_i[j]$ and the average predicted probability $s_i[j]$ over the dataset. Averaging these discrepancies across all $C$ classes yields a calibration loss that captures class-specific trends in over- and under-confidence. MDCA thus provides a more fine-grained and stable calibration signal, which is particularly beneficial in settings with many classes or imbalanced prediction distributions. For a minibatch of size $m$,

\begin{equation}
\begin{aligned}
    \mathcal{L}_{\mathrm{mdca}}
    &= \frac{1}{C} \sum_{j=1}^C 
    \Big\lvert \frac{1}{m} \sum_{i=1}^m c_i[j] 
    - \frac{1}{m} \sum_{i=1}^m s_i[j] \Big\lvert
\end{aligned}
\label{eq:mdca}
\end{equation}

Eq. \ref{eq:mdca} extends the DCA concept to the multi-class setting by considering each class $j = 1, \dots, C$. For each class, it computes the difference between the average ground-truth indicator $c_i[j]$ (which equals $1$ if sample $i$ belongs to class $j$ and $0$ otherwise) and the average predicted probability $s_i[j]$ across all $m$ samples in a minibatch. The loss is then averaged over all $C$ classes, providing a fine-grained calibration signal that captures errors in predicted probabilities across the full class distribution.

\noindent\textbf{Additional implementation details}. In this paragraph, we elaborate on the implementation details, providing additional information beyond what is included in the main paper. In backbone fine-tuning methods, we only fine-tune the LoRA-adapted layers of the transformer blocks, while keeping all original weights frozen. Specifically, for both the image and text encoders, only the LoRA matrices $W_{A}$ and $W_{B}$ of the query, key, and value projections in the attention modules are trainable. All other parameters, including projection, MLP, and layer normalization weights, remain frozen to reduce memory footprint and improve training efficiency. We keep the rank of LoRA metrices to 2 and a dropout rate of 0.25 to regularize the input of the LoRA modules. Furthermore, we set the number of bins to $G=10$ for all the calibration metrics.

\section{Additional empirical evaluation}
In this section, we 
provide additional experimental analyses and a detailed breakdown of the results presented in the main paper.

\begin{table*}[!t]
    \centering
    \caption{\textbf{Comparison of FL methods on the \textit{in-distribution} setting with additional calibration metrics} using datasets: CIFAR-10 and CIFAR-100. Best number is in \textbf{bold}, and second best is \underline{underlined}.Legends: \colorbox{amber!13}{prompt tuning methods}, and \colorbox{darkgreen!13}{backbone fine-tuning methods}.}
    \scalebox{0.75}{
    \begin{tabular}{l|cc|cc|cc}
         \toprule
         \textbf{Method}
         & \multicolumn{2}{c|}{\textbf{\makecell{FL setting \\(w/o calib)}}}
         & \multicolumn{2}{c|}{\textbf{\makecell{FL setting \\(w/ calib: DCA \cite{dca})}}}
         & \multicolumn{2}{c}{\textbf{\makecell{FL setting \\(w/ calib: MDCA \cite{hebbalaguppe2022stitch})}}} \\
         
         & Brier Score $\downarrow$ & NLL $\downarrow$
         & Brier Score $\downarrow$ & NLL $\downarrow$
         & Brier Score $\downarrow$ & NLL $\downarrow$ \\
         \midrule

         \rowcolor{gray!30}
         \multicolumn{7}{c}{CIFAR-10} \\

        \rowcolor{amber!13} PromptFL \cite{promptfl}
         & 0.12 & 0.23
         & 0.12 & 0.24
         & 0.12 & 0.23 \\

         \rowcolor{amber!13}FedPGP \cite{fedpgp}
         & 0.13 & 0.28
         & 0.13 & 0.28
         & 0.14 & 0.30 \\

         \rowcolor{amber!13}FedOTP \cite{fedotp}
         & 0.09 & 0.20
         & 0.09 & 0.20
         & 0.09 & 0.20 \\

         \rowcolor{amber!13}FedPHA \cite{fedpha}
         & 0.10 & 0.20
         & 0.10 & 0.20
         & 0.10 & 0.20 \\

         \midrule

         \rowcolor{darkgreen!13}AdaptFormer \cite{adaptformer}
         & 0.08 & 0.16
         & 0.08 & 0.15
         & 0.08 & 0.15 \\

         \rowcolor{darkgreen!13} LayerNorm \cite{layernorm} &0.11 &0.24 &0.10 & 0.22 &0.11 &0.23 \\

         \rowcolor{darkgreen!13}LoRA \cite{lora}
         & \textbf{0.05} & \textbf{0.09}
         & \textbf{0.05} & \textbf{0.09}
         & \textbf{0.05} & \textbf{0.09} \\

         \rowcolor{darkgreen!13}VeRA \cite{vera}
         & 0.10 & 0.21
         & 0.10 & 0.21
         & 0.10 & 0.22 \\

         \rowcolor{darkgreen!13}DoRA \cite{dora}
         & \underline{0.06} & \underline{0.11}
         & \underline{0.06} & \underline{0.10}
         & \underline{0.06} & \underline{0.10} \\

         \midrule
         \rowcolor{gray!30}
         \multicolumn{7}{c}{CIFAR-100} \\

        \rowcolor{amber!13} PromptFL \cite{promptfl}
         & 0.36 & 0.93
         & 0.36 & 0.92
         & 0.36 & 0.92 \\

         \rowcolor{amber!13}FedPGP \cite{fedpgp}
         & 0.42 & 1.12
         & 0.38 & 1.00
         & 0.38 & 0.98 \\

        \rowcolor{amber!13} FedOTP \cite{fedotp}
         & 0.38 & 1.00
         & 0.37 & 0.97
         & 0.38 & 1.00 \\

         \rowcolor{amber!13}FedPHA \cite{fedpha}
         & 0.35 & 0.91
         & 0.34 & 0.90
         & 0.34 & 0.91 \\

         \midrule

         \rowcolor{darkgreen!13}AdaptFormer \cite{adaptformer}
         & \underline{0.34} & \underline{0.89}
         & 0.30 & 0.72
         & 0.29 & 0.69 \\

         \rowcolor{darkgreen!13} LayerNorm \cite{layernorm} &0.41 &1.15 &0.40 &1.12 &0.43 &1.20 \\

         \rowcolor{darkgreen!13}LoRA \cite{lora}
         & 0.50 & 1.43
         & \textbf{0.23} & \textbf{0.52}
         & \textbf{0.23} & \textbf{0.52} \\

         \rowcolor{darkgreen!13}VeRA \cite{vera}
         & 0.39 & 1.03
         & 0.35 & 0.88
         & 0.35 & 0.90 \\

         \rowcolor{darkgreen!13}DoRA \cite{dora}
         & \textbf{0.33} & \textbf{0.86}
         & \underline{0.26} & \underline{0.60}
         & \underline{0.26} & \underline{0.59} \\

         \bottomrule
    \end{tabular}
    }
    \label{tab:nll}
\end{table*}

\subsection{Additional calibration metrics}
Following previous work on model calibration \cite{minderer2021revisiting}, we also report performance using two additional calibration metrics, Brier score and negative log-likelihood (NLL), which complement the three other metrics reported in the main paper. First, we describe the metrics, and then we present the results.

\noindent\textbf{Brier score} measures the mean squared error between predicted probabilities and one-hot labels. Given $|\mathcal{D}|$ examples and $C$ classes, with $i$ indexing examples and $c$ indexing classes, let $y(i,c)\in\{0,1\}$ denote the one-hot label (1 if $c=y(i)$, else 0) and $p(i,c)\in[0,1]$ the predicted probability with $\sum_{c=1}^{C} p(i,c)=1$. It can be defined as:
\begin{equation}
    \text{Brier score} = \frac{1}{|\mathcal{D}|}
    \sum_{i=1}^{|\mathcal{D}|} \sum_{c=1}^{C}
    \left( p_{i,c} - y_{i,c} \right)^{2},
\label{eq:brier}
\end{equation}
Lower Brier scores indicate better calibration, as the metric penalizes assigning probability mass to incorrect classes and rewards well-calibrated, appropriately uncertain predictions.

\noindent\textbf{Negative log-likelihood (NLL)} evaluates the average log probability assigned to the true class and strongly penalizes overconfident mistakes. Using the same notation above, with true class index $y$, it is defined as,
\begin{equation}
    \text{NLL} = -\frac{1}{|\mathcal{D}|}
    \sum_{i=1}^{|\mathcal{D}|}
    \log\!\left( p_{i,y} \right),
\label{eq:nll}
\end{equation}
Lower NLL corresponds to better calibration under the model's predictive distribution. As an unbounded, strictly proper scoring rule aligned with maximum-likelihood training, NLL is particularly sensitive to the tails of the confidence distribution, distinguishing models that differ mainly in how confidently they handle ambiguous samples.

In Table \ref{tab:nll}, we present the detailed Brier score and NLL calibration results of prompt tuning and backbone fine-tuning methods. We observe that backbone fine-tuning methods consistently achieve better calibration compared to the baselines. This is consistent with the findings in the main paper, where for all the calibration metrics, DoRA is better and/or competitive without needing extra regularization losses.

\begin{figure}[t]
    \centering
    \begin{subfigure}[t]{0.48\linewidth}
        \centering
        \includegraphics[width=\linewidth]{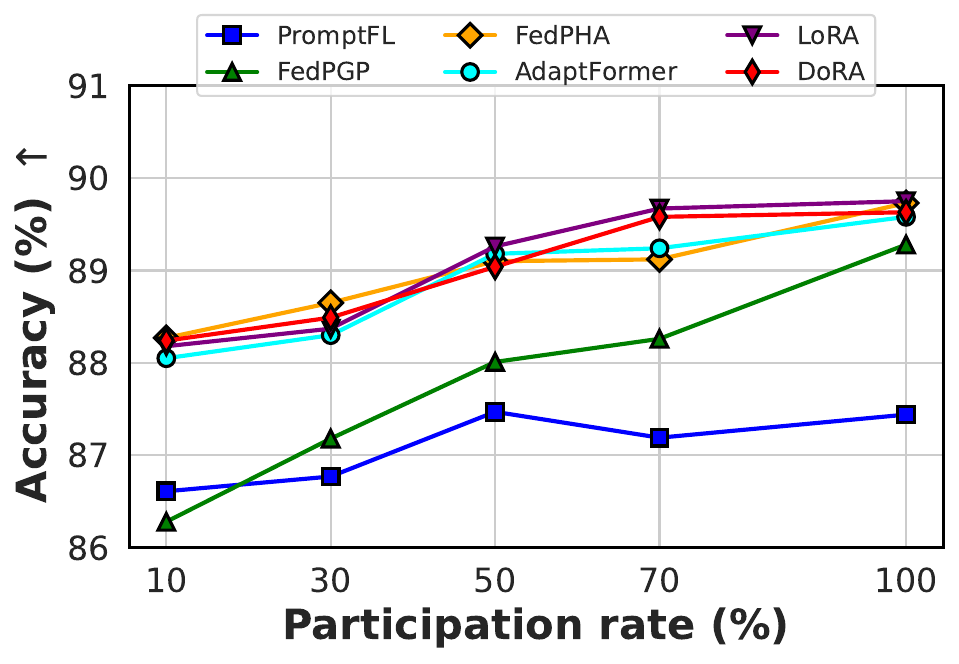}
        \caption{Accuracy}
        \label{fig:part_acc}
    \end{subfigure}
    \hfill
    \begin{subfigure}[t]{0.48\linewidth}
        \centering
        \includegraphics[width=\linewidth]{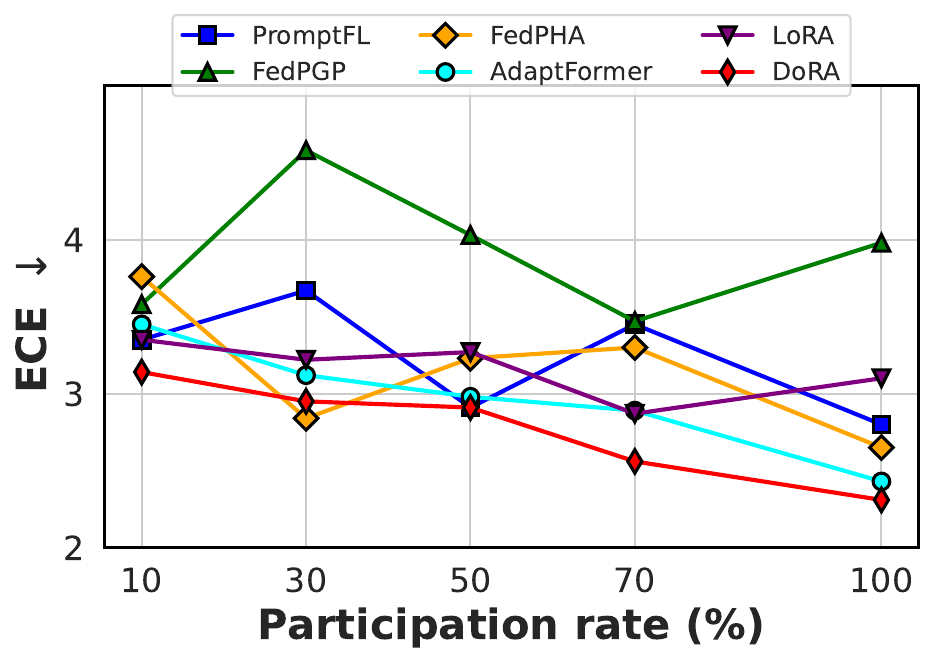}
        \caption{ECE}
        \label{fig:part_ece}
    \end{subfigure}
    \vspace{-0.2cm}
    \caption{\textbf{Effect of varying participation rates}. We report (a) Accuracy and (b) ECE metrics on OfficeHome dataset.}
    \label{fig:participations}
\end{figure}

\subsection{Impact of client participation rate}
Figure~\ref{fig:participations} evaluates client participation rates from 10\% to 100\% on OfficeHome. Accuracy generally increases with participation, although the trend is not monotonic for every method. FedPHA has the highest accuracy point estimate at 10\% and 30\% participation, with 88.27\% and 88.65\%, respectively, whereas LoRA has the highest point estimate from 50\% onward, reaching 89.75\% at full participation. Calibration is more method dependent. The ECE of AdaptFormer and DoRA decreases monotonically from 3.45\% to 2.43\% and from 3.14\% to 2.31\%, respectively, while the other methods fluctuate as participation changes. DoRA obtains the lowest ECE at 10\%, 70\%, and 100\%, ties PromptFL at 50\%, and is second to FedPHA at 30\%. Thus, increased participation generally benefits accuracy, but it does not guarantee a uniform calibration improvement.

\subsection{Effect of LoRA rank and encoder placement} 
Figure~\ref{fig:rank} separates the effects of LoRA rank $r$ and encoder placement on CIFAR-100 without calibration regularization. Increasing the rank from 2 to 16 does not yield a monotonic improvement in either accuracy or ECE; instead, performance varies only modestly within each placement. Encoder placement has a substantially larger effect. Applying LoRA to the image encoder or to both encoders outperforms the prompt-tuning references in both accuracy and ECE at every evaluated rank. LoRA-T i.e. LoRA tuning applied only on the text encoder, yields ECE close to the FedPHA reference, but its accuracy remains below FedOTP. These results show that rank 2 already provides a strong accuracy--calibration trade-off for image-only and joint adaptation, while increasing the rank adds trainable parameters without a consistent empirical benefit.

\begin{figure}[t]
    \centering
    \begin{subfigure}[t]{0.49\linewidth}
        \centering
        \includegraphics[width=\linewidth]{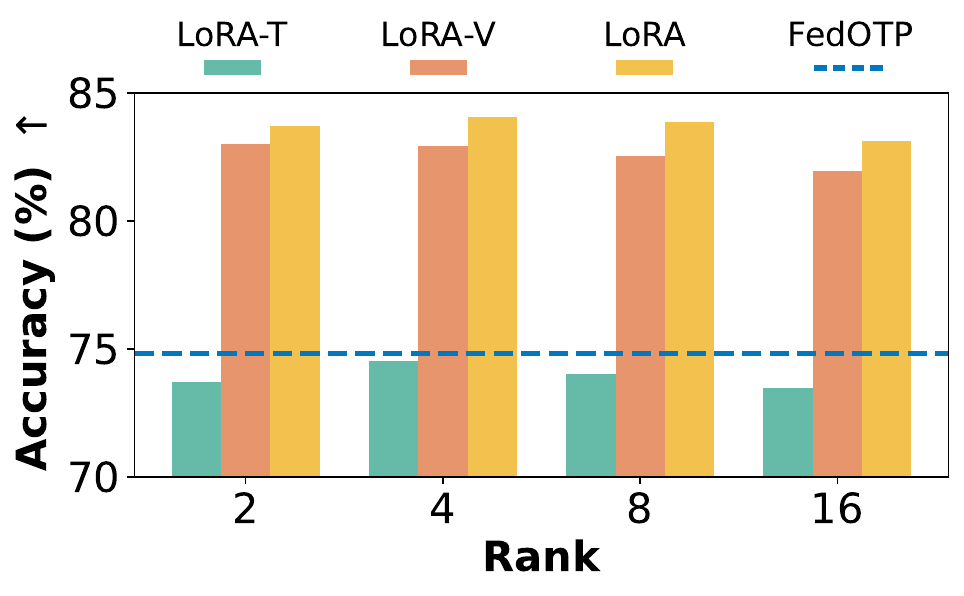}
    \end{subfigure}
    \hfill
    \begin{subfigure}[t]{0.49\linewidth}
        \centering
        \includegraphics[width=\linewidth]{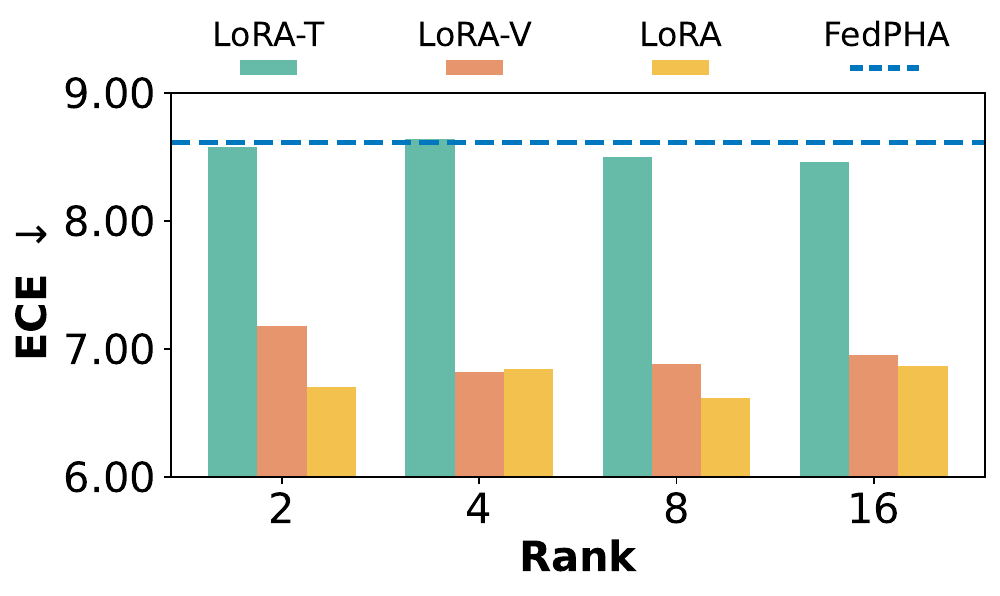}
    \end{subfigure}
    \caption{\textbf{Ablation on the LoRA rank and placement.} (left) Accuracy and (right) ECE on CIFAR-100 for different LoRA ranks $r$ and encoder placements. `T' and `V' denote applying LoRA to the text and vision encoders, respectively.}
    \label{fig:rank}
\end{figure}

\subsection{Trade-off between computational complexity and calibration}
Table~\ref{tab:lora_variants} compares accuracy and calibration against computational complexity, using the number of trainable parameters as a proxy, on CIFAR-10 without calibration regularization. We evaluate each backbone fine-tuning method under text-only, image-only, and joint text--image adaptation. Across most methods, adapting the image encoder produces substantially better accuracy and calibration than adapting only the text encoder. Jointly adapting both encoders, however, does not necessarily provide a further benefit despite its larger parameter budget. For example, LoRA improves from 92.64\% accuracy and 4.99\% ECE with text-only adaptation to 97.08\% accuracy and 2.94\% ECE with image-only adaptation. Updating both encoders increases its trainable parameters from 110.6K to 184.3K, yet yields a similar 97.04\% accuracy and 2.99\% ECE. DoRA exhibits the same pattern in accuracy and ECE, although its joint configuration improves MCE and ACE.

\begin{table*}[!t]
    \centering
    \caption{\textbf{Trade-off between computational complexity and calibration among backbone fine-tuning methods} using CIFAR-10 dataset. Best number is in \textbf{bold}, and second best is \underline{underlined}.}
    \scalebox{0.65}{
    \begin{tabular}{l|c|cccc|c|cccc|c|cccc}
         \toprule
         \textbf{Method} & \multicolumn{5}{c|}{\textbf{\makecell{Text-encoder}}} & \multicolumn{5}{c|}{\textbf{\makecell{Image-encoder}}} & \multicolumn{5}{c}{\textbf{\makecell{Both encoders}}} \\
         & \makecell{\#Tr. \\Params} & Acc. $\uparrow$ &ECE $\downarrow$ &MCE $\downarrow$ &ACE $\downarrow$ & \makecell{\#Tr. \\Params} & Acc. $\uparrow$ &ECE $\downarrow$ &MCE $\downarrow$ &ACE $\downarrow$ & \makecell{\#Tr. \\Params} & Acc. $\uparrow$ &ECE $\downarrow$ &MCE $\downarrow$ &ACE $\downarrow$\\
         \midrule
         CLIP-FT \cite{radford2019language} &38.1M &92.31	&5.16	&1.66	&4.34 &86.2M &95.57	&3.87	&1.41	&3.33 &124.3M &95.76	&3.44	&1.25	&2.48 \\

         AdaptFormer \cite{adaptformer} &30.8K	&92.51	&5.39	&1.87	&3.92 &46.1K &96.64	&3.06	&1.20	&2.09 &76.9K &96.80	&3.00	&1.16	&1.93 \\

         LayerNorm \cite{layernorm} &49.1K &90.42	&5.56	&1.76	&4.70
& 75.3K &96.04	&3.45	&1.26	&2.39
& 124.4K &96.03	&3.49	&1.30	&2.21 \\

         LoRA \cite{lora} &73.7K &92.64	&4.99	&1.68	&3.77 &110.6K &97.08	&\textbf{2.94}	&1.14	&1.98 &184.3K &97.04	&\underline{2.99}	&\underline{1.12}	&2.06 \\

    VeRA \cite{vera} &36.9K	&88.50	&8.54	&2.34	&7.91 &55.3K &88.77	&8.55	&2.37	&8.19 &92.2K &88.93	&8.30	&2.25	&7.82 \\

    DoRA \cite{dora} &92.2K &92.65	&5.08	&1.64	&4.06 &138.2K &\textbf{97.23}	&\underline{2.99}	&\underline{1.12}	&\underline{1.90} &230.4K &\underline{97.19}	&3.05	&\textbf{1.08}	&\textbf{1.85} \\

         \bottomrule
    \end{tabular}
    }
    \label{tab:lora_variants}
\end{table*}

\subsection{Proof of Lemma 6.1: Federated--centralized drift}
The proof is direct. For $K=1$, Eq.~\eqref{eq:fed_central_drift} is zero: one FedAvg step is identical to one centralized gradient step. For $K>1$, non-IID gradients lead the clients along different optimization trajectories, producing a federated--centralized parameter gap.

\subsection{Proof of Proposition 6.2 : Federated–centralized prediction stability}
By the mean-value inequality,
\[
\|\mathbf z(\vect{x};\theta_{t+1}^{G})
-\mathbf z(\vect{x};\theta_{t,K}^{C})\|_2
\leq
K_z(\vect{x})
\|\theta_{t+1}^{G}-\theta_{t,K}^{C}\|_2.
\]
Applying the Lipschitz continuity of the softmax gives Eq.10.

\subsection{Empirical Validation of the Jacobian-Weighted Drift Condition}
Proposition 6.2 shows that the effect of the federated--centralized parameter gap on the predictions is controlled by the product of the logit-Jacobian sensitivity and the parameter drift. We empirically evaluate this quantity on CIFAR-10 without calibration regularization, using LoRA as the backbone fine-tuning method (BFT) and FedPHA as the prompt-tuning (PT) baseline. For LoRA, the parameter vector contains the trainable LoRA factors; for FedPHA, it contains the shared prompt parameters, while the personalized prompts are held fixed. At each selected communication round $t$, the federated and centralized branches start from the same parameters $\theta_t$. The federated branch performs the prescribed local updates followed by FedAvg, producing $\theta_{t+1}^{\mathcal G}$, while a matched centralized branch performs the same number of updates using the same participating clients and their renormalized sample weights, producing $\theta_{t,K}^{\mathcal C}$. For method $M\in\{\mathrm{LoRA},\mathrm{FedPHA}\}$, we define
\begin{equation}
    \Delta_{M,t}^{\mathcal G,\mathcal C}
    =
    \left\|\Delta_{M,t}^{\mathcal G,\mathcal C}\right\|_2.
    \label{eq:empirical_jacobian_drift}
\end{equation}
A lower value of $\widehat S_M(t)$ corresponds to a smaller estimated stability term. Provided that the interpolation grid approximates the segment supremum sufficiently well, observing $\widehat S_{\mathrm{LoRA}}(t)<\widehat S_{\mathrm{FedPHA}}(t)$ across communication rounds would support the sufficient condition identified in Proposition 6.2. This comparison would constitute empirical evidence for the proposed mechanism, rather than a universal proof: a complete evaluation should report uncertainty across random seeds and the fraction of probe images for which the pointwise inequality holds. The curves currently shown in Fig.~\ref{fig:jacobian_drift} illustrate the expected diagnostic and must be replaced by measurements from the matched federated and centralized runs before being interpreted as empirical evidence.
A lower value of $\widehat S_M(t)$ corresponds to a smaller estimated stability term. As shown in Fig.~\ref{fig:jacobian_drift}, LoRA produces a consistently smaller Jacobian-weighted federated drift than FedPHA at every evaluated communication round. The separation is already visible in the first round and increases throughout training: by round 50, the FedPHA value is approximately $0.040$, whereas the LoRA value remains below $0.010$. Therefore, the measured average satisfies
\begin{equation}
    \widehat S_{\mathrm{LoRA}}(t)
    <
    \widehat S_{\mathrm{FedPHA}}(t)
    \qquad
    \text{for every evaluated round }t.
    \label{eq:empirical_lora_fedpha_ordering}
\end{equation}
This ordering is precisely the condition under which Proposition 6.2 gives LoRA a tighter estimated upper bound on the federated--centralized prediction gap. It is also consistent with the CIFAR-10 calibration results in Tab.3 of the main paper, where LoRA achieves an ECE of 2.99\%, compared with 4.59\% for FedPHA, without calibration regularization. Thus, Fig.~\ref{fig:jacobian_drift} provides empirical support for the proposed stability mechanism: LoRA's smaller Jacobian-weighted federated drift is associated with a smaller calibration error in this setting. This result validates the ordering for the averaged diagnostic, evaluated checkpoints, and CIFAR-10 configuration; it does not establish a universal or pointwise inequality for every input and dataset.

\begin{figure}[t]
    \centering
    \includegraphics[width=\linewidth]{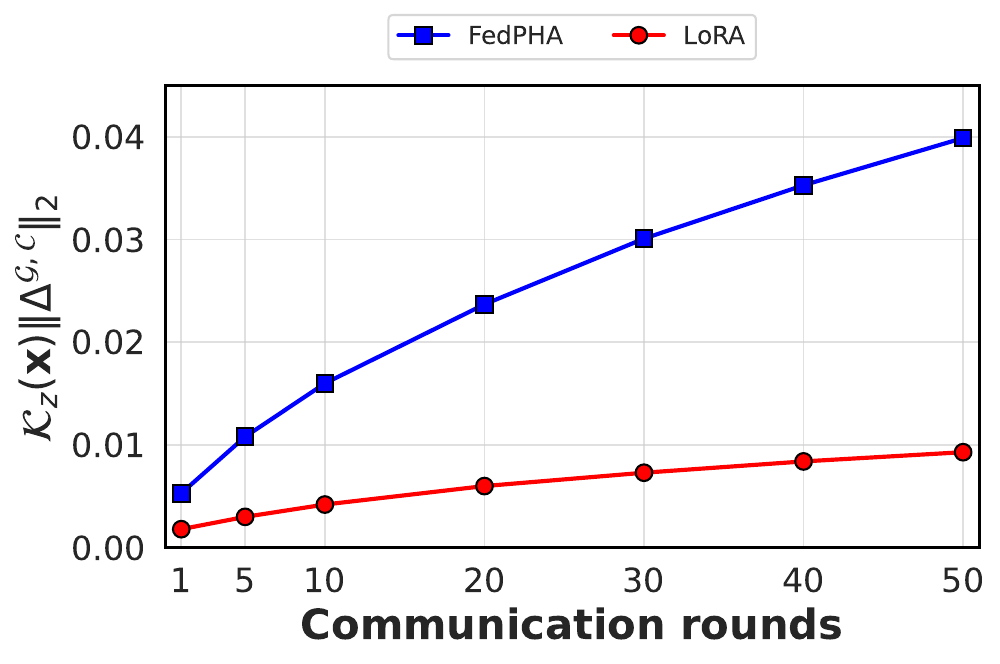}
\caption{\textbf{Jacobian-weighted federated drift across communication rounds.}
Using CIFAR-10 dataset without calibration regularization, we report the probe-set average of
$\widehat{\mathcal{K}}_{z,t}^{M}(\vect{x})
\|\Delta_{M,t}^{\mathcal{G},\mathcal{C}}\|_2$
for LoRA as BFT method and FedPHA as PT method. LoRA exhibits consistently lower Jacobian-weighted drift, indicating a tighter estimated bound on the federated--centralized prediction gap.}
    \label{fig:jacobian_drift}
\end{figure}

\subsection{Detailed results of the main tables} 
In Table \ref{tab:pacs}, \ref{tab:officehome} and \ref{tab:vlcs}, we provide the detailed results of individual datasets PACS, OfficeHome and VLCS respectively in the domain-generalization setting. The average results of them are reported in the Table. 4 of the main paper. Similarly in Table \ref{tab:food101}, \ref{tab:dtd}, \ref{tab:caltech101}, \ref{tab:flowers102} and \ref{tab:oxfordpets}, we provide the detailed results of Food101, DTD, Caltech101, Flowers102, and OxfordPets datasets, corresponding to Table 5 of the main paper, for the base-to-new generalization setting.

\begin{table*}[!t]
    \centering
    \small
    \caption{\textbf{Comparison of FL methods on the \textit{domain generalization} setting} using the PACS dataset. Best number is in \textbf{bold}, and second best is \underline{underlined}.Legends: \colorbox{amber!13}{prompt tuning methods}, and \colorbox{darkgreen!13}{backbone fine-tuning methods}.}
    \scalebox{0.76}{
\begin{tabular}{l|cccc|cccc|cccc}
        \toprule
        & \multicolumn{4}{c|}{\textbf{Art Painting}}
        & \multicolumn{4}{c|}{\textbf{Cartoon}}
        & \multicolumn{4}{c}{\textbf{Photo}} \\
        \cmidrule(lr){2-5}
        \cmidrule(lr){6-9}
        \cmidrule(lr){10-13}

        \multirow{-2}{*}{\textbf{Method}}
        & Acc. $\uparrow$ & ECE $\downarrow$ & MCE $\downarrow$ & ACE $\downarrow$
        & Acc. $\uparrow$ & ECE $\downarrow$ & MCE $\downarrow$ & ACE $\downarrow$
        & Acc. $\uparrow$ & ECE $\downarrow$ & MCE $\downarrow$ & ACE $\downarrow$ \\
        \midrule

        \rowcolor{gray!30}
        \multicolumn{13}{c}{FL setting (\textit{without calibration})} \\

        \rowcolor{amber!13}
        PromptFL \cite{promptfl}
        & 98.83 & 0.95 & 0.40 & 0.48
        & 99.20 & 0.98 & 0.42 & \textbf{0.31}
        & 99.89 & 0.55 & 0.24 & 0.38 \\

        \rowcolor{amber!13}
        FedPGP \cite{fedpgp}
        & 99.02 & 0.95 & 0.40 & 0.61
        & 99.66 & 0.76 & 0.31 & 0.55
        & \textbf{99.95} & 0.37 & \textbf{0.13} & 0.24 \\

        \rowcolor{amber!13}
        FedOTP \cite{fedotp}
        & 98.34 & 4.04 & 1.62 & 3.98
        & 99.04 & 3.68 & 1.55 & 3.58
        & 99.80 & 2.22 & 1.38 & 2.15 \\

        \rowcolor{amber!13}
        FedPHA \cite{fedpha}
        & 99.02 & 0.98 & 0.26 & 0.44
        & 99.82 & 0.76 & 0.34 & 0.53
        & 99.90 & \textbf{0.31} & 0.27 & 0.31 \\

        \midrule

        \rowcolor{darkgreen!13}AdaptFormer \cite{adaptformer}
        & 99.28 & 0.68 & 0.24 & 0.42
        & 99.87 & 0.71 & 0.30 & 0.39
        & 99.91 & 0.40 & 0.18 & 0.27 \\
\end{tabular}
    }
    \label{tab:pacs}
\end{table*}
\begin{table*}[!t]
    \centering
    \small
    \caption{\textbf{Comparison of FL methods on the \textit{domain generalization} setting} using the OfficeHome dataset. Best number is in \textbf{bold}, and second best is \underline{underlined}.Legends: \colorbox{amber!13}{prompt tuning methods}, and \colorbox{darkgreen!13}{backbone fine-tuning methods}.}
    \scalebox{0.76}{
    \begin{tabular}{l|cccc|cccc|cccc}
        \toprule
        & \multicolumn{4}{c|}{\textbf{Art}}
        & \multicolumn{4}{c|}{\textbf{Clipart}}
        & \multicolumn{4}{c}{\textbf{Product}} \\
        \cmidrule(lr){2-5} \cmidrule(lr){6-9} \cmidrule(lr){10-13}
        
        \multirow{-2}{*}{\textbf{Method}}
        & Acc. $\uparrow$ & ECE $\downarrow$ & MCE $\downarrow$ & ACE $\downarrow$
        & Acc. $\uparrow$ & ECE $\downarrow$ & MCE $\downarrow$ & ACE $\downarrow$
        & Acc. $\uparrow$ & ECE $\downarrow$ & MCE $\downarrow$ & ACE $\downarrow$ \\

        \midrule
        \rowcolor{gray!30}
        \multicolumn{13}{c}{FL setting (\textit{without calibration})} \\

        \rowcolor{amber!13}PromptFL \cite{promptfl}
        & 87.55 & 3.83 & 1.00 & 3.96
        & 75.35 & 3.57 & 1.13 & 3.27
        & 93.85 & 2.00 & 0.60 & 1.51 \\

        \rowcolor{amber!13}FedPGP \cite{fedpgp}
        & 89.10 & 5.32 & 1.24 & 3.74
        & 78.55 & 4.24 & 1.21 & 3.19
        & 95.50 & 2.84 & 0.71 & 1.78 \\

        \rowcolor{amber!13}FedOTP \cite{fedotp}
        & 81.95 & 23.56 & 5.48 & 23.43
        & 77.15 & 17.57 & 3.25 & 17.53
        & 92.40 & 21.04 & 3.45 & 21.00 \\

        \rowcolor{amber!13}FedPHA \cite{fedpha}
        & 88.05 & 3.36 & \underline{0.77} & 2.60
        & \underline{81.35} & 3.07 & \underline{0.76} & 2.88
        & \textbf{96.00} & 1.85 & 0.57 & \underline{1.09} \\

        \midrule

        \rowcolor{darkgreen!13}AdaptFormer \cite{adaptformer}
        & 88.85 & \underline{3.25} & 0.79 & 2.35
        & 80.95 & \underline{3.00} & 0.78 & \underline{2.67}
        & 95.20 & \underline{1.74} & \underline{0.54} & 1.10 \\
\end{tabular}
    }
    \label{tab:officehome}
\end{table*}
\begin{table*}[!t]
    \centering
    \small
    \caption{\textbf{Comparison of FL methods on the \textit{domain generalization} setting} using the VLCS dataset. Best number is in \textbf{bold}, and second best is \underline{underlined}.Legends: \colorbox{amber!13}{prompt tuning methods}, and \colorbox{darkgreen!13}{backbone fine-tuning methods}.}
    \scalebox{0.76}{
    \begin{tabular}{l|cccc|cccc|cccc}
        \toprule
        & \multicolumn{4}{c|}{\textbf{Caltech}} 
        & \multicolumn{4}{c|}{\textbf{Labelme}} 
        & \multicolumn{4}{c}{\textbf{Pascal-VOC}} \\
        \cmidrule(lr){2-5} \cmidrule(lr){6-9} \cmidrule(lr){10-13}
        
        \multirow{-2}{*}{\textbf{Method}} 
        & Acc. $\uparrow$ & ECE $\downarrow$ & MCE $\downarrow$ & ACE $\downarrow$ 
        & Acc. $\uparrow$ & ECE $\downarrow$ & MCE $\downarrow$ & ACE $\downarrow$ 
        & Acc. $\uparrow$ & ECE $\downarrow$ & MCE $\downarrow$ & ACE $\downarrow$ \\ 

        \midrule
        \rowcolor{gray!30}
        \multicolumn{13}{c}{FL setting (\textit{without calibration})} \\

        \rowcolor{amber!13}PromptFL \cite{promptfl}
        & \textbf{100.00} & 2.22 & 1.83 & 2.15
        & 78.24 & 10.11 & 4.84 & 11.67
        & 89.27 & 2.70 & 1.19 & 2.58 \\

        \rowcolor{amber!13}FedPGP \cite{fedpgp}
        & \textbf{100.00} & 2.30 & 1.23 & 2.25
        & 81.37 & 4.87 & 2.09 & 5.68
        & 90.77 & 2.06 & 1.07 & 2.30 \\

        \rowcolor{amber!13}FedOTP \cite{fedotp}
        & \textbf{100.00} & 2.80 & 1.82 & 2.80
        & 81.96 & 6.20 & 2.13 & 6.08
        & 90.58 & 6.21 & 1.80 & 5.59 \\

        \rowcolor{amber!13}FedPHA \cite{fedpha}
        & \textbf{100.00} & 2.51 & 1.18 & 2.39
        & 81.98 & 4.85 & 1.92 & 5.40
        & 90.67 & 2.11 & 0.88 & 1.82 \\

        \midrule

        \rowcolor{darkgreen!13} AdaptFormer \cite{adaptformer}
        & \textbf{100.00} & \underline{1.95} & \underline{1.10} & \underline{1.92}
        & \underline{82.30} & \underline{4.55} & \underline{1.75} & \underline{4.95}
        & 90.95 & \underline{1.98} & \underline{0.82} & \underline{1.75} \\
\end{tabular}
    }
    \label{tab:vlcs}
\end{table*}
\begin{table*}[!t]
    \centering
    \small
    \caption{\textbf{Comparison of FL methods on the \textit{base-to-new generalization} setting} using the Food101 dataset. Best number is in \textbf{bold}, and second best is \underline{underlined}.Legends: \colorbox{amber!13}{prompt tuning methods}, and \colorbox{darkgreen!13}{backbone fine-tuning methods}.}
    \scalebox{0.8}{
    \begin{tabular}{l|ccc|ccc|ccc|ccc}
        \toprule
        & \multicolumn{3}{c|}{\textbf{Acc. $\uparrow$}}
        & \multicolumn{3}{c|}{\textbf{ECE $\downarrow$}}
        & \multicolumn{3}{c|}{\textbf{MCE $\downarrow$}}
        & \multicolumn{3}{c}{\textbf{ACE $\downarrow$}} \\
        \cmidrule(lr){2-4}
        \cmidrule(lr){5-7}
        \cmidrule(lr){8-10}
        \cmidrule(lr){11-13}
        
        \multirow{-2}{*}{\textbf{Method}}
        & Base & New & HM
        & Base & New & HM
        & Base & New & HM
        & Base & New & HM \\

        \midrule
        \rowcolor{gray!30}
        \multicolumn{13}{c}{FL setting (\textit{without calibration})} \\

        \rowcolor{amber!13}PromptFL \cite{promptfl}
        & 89.78 & \underline{90.79} & 90.28
        & \textbf{0.48} & 0.70 & \textbf{0.57}
        & \textbf{0.13} & \textbf{0.17} & \textbf{0.15}
        & \textbf{0.39} & 0.61 & \textbf{0.47} \\

        \rowcolor{amber!13}FedPGP \cite{fedpgp}
        & 87.54 & 87.94 & 87.74
        & 1.19 & 1.08 & 1.13
        & 0.38 & 0.35 & 0.37
        & 1.16 & 1.02 & 1.09 \\

        \rowcolor{amber!13}FedOTP \cite{fedotp}
        & 15.18 & 40.18 & 22.03
        & 39.93 & 17.49 & 24.32
        & 8.05 & 4.02 & 5.36
        & 39.93 & 17.48 & 24.31 \\

        \rowcolor{amber!13}FedPHA \cite{fedpha}
        & 58.11 & 70.69 & 63.79
        & 19.21 & 7.56 & 10.85
        & 7.37 & 2.78 & 4.04
        & 18.19 & 7.31 & 10.43 \\

        \midrule

        \rowcolor{darkgreen!13}AdaptFormer \cite{adaptformer}
        & \underline{89.95} & \textbf{90.92} & \textbf{90.43}
        & 0.55 & \textbf{0.68} & 0.61
        & 0.15 & 0.20 & 0.17
        & 0.42 & \underline{0.58} & 0.49 \\
\bottomrule
    \end{tabular}
    }
    \label{tab:food101}
\end{table*}
\begin{table*}[!t]
    \centering
    \small
    \caption{\textbf{Comparison of FL methods on the \textit{base-to-new generalization} setting} using the DTD dataset. Best number is in \textbf{bold}, and second best is \underline{underlined}.Legends: \colorbox{amber!13}{prompt tuning methods}, and \colorbox{darkgreen!13}{backbone fine-tuning methods}.}
    \scalebox{0.8}{
    \begin{tabular}{l|ccc|ccc|ccc|ccc}
        \toprule
        & \multicolumn{3}{c|}{\textbf{Acc. $\uparrow$}}
        & \multicolumn{3}{c|}{\textbf{ECE $\downarrow$}}
        & \multicolumn{3}{c|}{\textbf{MCE $\downarrow$}}
        & \multicolumn{3}{c}{\textbf{ACE $\downarrow$}} \\
        
        \cmidrule(lr){2-4}
        \cmidrule(lr){5-7}
        \cmidrule(lr){8-10}
        \cmidrule(lr){11-13}
        
        \multirow{-2}{*}{\textbf{Method}}
        & Base & New & HM
        & Base & New & HM
        & Base & New & HM
        & Base & New & HM \\

        \midrule
        \rowcolor{gray!30}
        \multicolumn{13}{c}{FL setting (\textit{without calibration})} \\

        \rowcolor{amber!13}PromptFL \cite{promptfl}
        & 76.85 & 44.93 & 56.71
        & \textbf{6.72} & 24.65 & \textbf{10.56}
        & \textbf{2.32} & 7.38 & \textbf{3.54}
        & \textbf{6.35} & 24.65 & 10.10 \\

        \rowcolor{amber!13}FedPGP \cite{fedpgp}
        & 67.88 & 50.87 & 58.16
        & 13.98 & 19.19 & 16.18
        & 5.88 & \underline{4.27} & 4.95
        & 13.88 & 19.12 & 16.08 \\

        \rowcolor{amber!13}FedOTP \cite{fedotp}
        & 27.65 & 42.17 & 33.40
        & 41.26 & \textbf{13.68} & 20.55
        & 11.00 & \textbf{2.95} & \underline{4.66}
        & 41.04 & 13.71 & 20.55 \\

        \rowcolor{amber!13}FedPHA \cite{fedpha}
        & 27.87 & 38.29 & 32.26
        & 50.52 & 20.53 & 29.20
        & 25.81 & 4.50 & 7.67
        & 48.96 & 20.50 & 28.90 \\

        \midrule

        \rowcolor{darkgreen!13}AdaptFormer \cite{adaptformer}
        & 79.80 & 54.69 & 64.90
        & 8.00 & 18.67 & 11.20
        & 3.50 & 7.15 & 4.70
        & 8.10 & \textbf{12.09} & \underline{9.70} \\
\bottomrule
    \end{tabular}
    }
    \label{tab:dtd}
\end{table*}
\begin{table*}[!t]
    \centering
    \small
    \caption{\textbf{Comparison of FL methods on the \textit{base-to-new generalization} setting} using the Caltech101 dataset. Best number is in \textbf{bold}, and second best is \underline{underlined}.Legends: \colorbox{amber!13}{prompt tuning methods}, and \colorbox{darkgreen!13}{backbone fine-tuning methods}.}
    \scalebox{0.8}{
    \begin{tabular}{l|ccc|ccc|ccc|ccc}
        \toprule
        & \multicolumn{3}{c|}{\textbf{Acc. $\uparrow$}}
        & \multicolumn{3}{c|}{\textbf{ECE $\downarrow$}}
        & \multicolumn{3}{c|}{\textbf{MCE $\downarrow$}}
        & \multicolumn{3}{c}{\textbf{ACE $\downarrow$}} \\
        \cmidrule(lr){2-4}
        \cmidrule(lr){5-7}
        \cmidrule(lr){8-10}
        \cmidrule(lr){11-13}
        
        \multirow{-2}{*}{\textbf{Method}}
        & Base & New & HM
        & Base & New & HM
        & Base & New & HM
        & Base & New & HM \\

        \midrule
        \rowcolor{gray!30}
        \multicolumn{13}{c}{FL setting (\textit{without calibration})} \\

        \rowcolor{amber!13}PromptFL \cite{promptfl}
        & 97.55 & 92.36 & 94.88
        & \textbf{0.77} & 4.41 & \textbf{1.31}
        & \textbf{0.21} & 3.77 & \textbf{0.40}
        & \textbf{0.36} & 4.89 & \textbf{0.67} \\

        \rowcolor{amber!13}FedPGP \cite{fedpgp}
        & 96.36 & 92.43 & 94.36
        & 1.39 & 4.08 & 2.07
        & 0.62 & 3.01 & 1.03
        & 0.87 & 3.64 & 1.40 \\

        \rowcolor{amber!13}FedOTP \cite{fedotp}
        & 38.97 & 67.98 & 49.54
        & 18.51 & 11.40 & 14.11
        & 3.68 & 2.61 & 3.05
        & 18.68 & 11.31 & 14.09 \\

        \rowcolor{amber!13}FedPHA \cite{fedpha}
        & 82.15 & 87.03 & 84.52
        & 5.42 & 5.55 & 5.48
        & 1.95 & 2.75 & 2.28
        & 5.16 & 4.75 & 4.95 \\

        \midrule

        \rowcolor{darkgreen!13}AdaptFormer \cite{adaptformer}
        & 98.20 & \underline{93.55} & \underline{95.82}
        & 1.05 & \underline{2.80} & 1.53
        & 0.36 & \underline{0.95} & 0.52
        & 0.70 & \underline{2.20} & 1.06 \\
\bottomrule
    \end{tabular}
    }
    \label{tab:caltech101}
\end{table*}
\begin{table*}[!t]
    \centering
    \small
    \caption{\textbf{Comparison of FL methods on the \textit{base-to-new generalization} setting} using the Flowers102 dataset. Best number is in \textbf{bold}, and second best is \underline{underlined}.Legends: \colorbox{amber!13}{prompt tuning methods}, and \colorbox{darkgreen!13}{backbone fine-tuning methods}.}
    \scalebox{0.8}{
    \begin{tabular}{l|ccc|ccc|ccc|ccc}
        \toprule
        & \multicolumn{3}{c|}{\textbf{Acc. $\uparrow$}}
        & \multicolumn{3}{c|}{\textbf{ECE $\downarrow$}}
        & \multicolumn{3}{c|}{\textbf{MCE $\downarrow$}}
        & \multicolumn{3}{c}{\textbf{ACE $\downarrow$}} \\
        \cmidrule(lr){2-4}
        \cmidrule(lr){5-7}
        \cmidrule(lr){8-10}
        \cmidrule(lr){11-13}
        
        \multirow{-2}{*}{\textbf{Method}}
        & Base & New & HM
        & Base & New & HM
        & Base & New & HM
        & Base & New & HM \\

        \midrule
        \rowcolor{gray!30}
        \multicolumn{13}{c}{FL setting (\textit{without calibration})} \\

        \rowcolor{amber!13}PromptFL \cite{promptfl}
        & 95.82 & 64.26 & 76.93
        & \underline{3.05} & 15.74 & 5.11
        & 1.04 & 3.89 & 1.64
        & 2.91 & 15.74 & 4.91 \\

        \rowcolor{amber!13}FedPGP \cite{fedpgp}
        & 73.79 & 65.21 & 69.24
        & 7.53 & 10.88 & 8.90
        & 3.24 & \underline{2.66} & 2.92
        & 7.41 & 10.86 & 8.81 \\

       \rowcolor{amber!13} FedOTP \cite{fedotp}
        & 15.15 & 32.85 & 20.73
        & 46.74 & 13.96 & 21.50
        & 9.81 & 3.28 & 4.91
        & 46.72 & 14.09 & 21.65 \\

        \rowcolor{amber!13}FedPHA \cite{fedpha}
        & 29.43 & 49.37 & 36.88
        & 48.27 & 16.31 & 24.38
        & 21.26 & 3.57 & 6.11
        & 45.80 & 16.28 & 24.02 \\

        \midrule

        \rowcolor{darkgreen!13}AdaptFormer \cite{adaptformer}
        & 97.20 & \textbf{72.56} & \underline{83.09}
        & 4.00 & \underline{10.48} & 5.79
        & \underline{1.00} & \textbf{2.39} & 1.41
        & 3.10 & \underline{7.12} & 4.32 \\
\bottomrule
    \end{tabular}
    }
    \label{tab:flowers102}
\end{table*}
\begin{table*}[!t]
    \centering
    \small
    \caption{\textbf{Comparison of FL methods on the \textit{base-to-new generalization} setting} using the OxfordPets dataset. Best number is in \textbf{bold}, and second best is \underline{underlined}.Legends: \colorbox{amber!13}{prompt tuning methods}, and \colorbox{darkgreen!13}{backbone fine-tuning methods}.}
    \scalebox{0.8}{
    \begin{tabular}{l|ccc|ccc|ccc|ccc}
        \toprule
        & \multicolumn{3}{c|}{\textbf{Acc. $\uparrow$}}
        & \multicolumn{3}{c|}{\textbf{ECE $\downarrow$}}
        & \multicolumn{3}{c|}{\textbf{MCE $\downarrow$}}
        & \multicolumn{3}{c}{\textbf{ACE $\downarrow$}} \\
        \cmidrule(lr){2-4}
        \cmidrule(lr){5-7}
        \cmidrule(lr){8-10}
        \cmidrule(lr){11-13}
        
        \multirow{-2}{*}{\textbf{Method}}
        & Base & New & HM
        & Base & New & HM
        & Base & New & HM
        & Base & New & HM \\

        \midrule
        \rowcolor{gray!30}
        \multicolumn{13}{c}{FL setting (\textit{without calibration})} \\

        \rowcolor{amber!13}PromptFL \cite{promptfl}
        & \underline{95.48} & 96.76 & 96.12
        & 2.85 & 3.13 & 2.98
        & 0.79 & 0.91 & 0.85
        & 2.78 & 3.13 & 2.95 \\

        \rowcolor{amber!13}FedPGP \cite{fedpgp}
        & 94.86 & 97.12 & 95.98
        & 2.08 & 3.37 & 2.57
        & 0.86 & 1.42 & 1.07
        & 2.00 & 3.25 & 2.48 \\

        \rowcolor{amber!13}FedOTP \cite{fedotp}
        & 13.86 & 52.07 & 21.89
        & 64.69 & 18.64 & 28.94
        & 26.50 & 4.13 & 7.15
        & 64.55 & 18.63 & 28.91 \\

        \rowcolor{amber!13}FedPHA \cite{fedpha}
        & 44.75 & 80.04 & 57.40
        & 36.96 & 5.96 & 10.26
        & 20.54 & 1.69 & 3.12
        & 33.76 & 5.96 & 10.13 \\

        \midrule

        \rowcolor{darkgreen!13}AdaptFormer \cite{adaptformer}
        & 95.45 & \underline{97.50} & \underline{96.46}
        & \underline{0.56} & \underline{1.02} & \underline{0.72}
        & \underline{0.31} & \underline{0.39} & \underline{0.35}
        & \underline{0.61} & \underline{0.65} & \underline{0.63} \\
\bottomrule
    \end{tabular}
    }
    \label{tab:oxfordpets}
\end{table*}

\end{document}